\documentclass[lettersize,journal]{IEEEtran}
\usepackage{amsmath,amsfonts}
\usepackage{array}
\usepackage[caption=false,font=normalsize,labelfont=sf,textfont=sf]{subfig}
\usepackage{textcomp}
\usepackage{stfloats}
\usepackage{url}
\usepackage{verbatim}
\usepackage{graphicx}
\def\BibTeX{{\rm B\kern-.05em{\sc i\kern-.025em b}\kern-.08em
    T\kern-.1667em\lower.7ex\hbox{E}\kern-.125emX}}
\usepackage{balance}

\usepackage{amsmath}
\usepackage{amssymb}
\usepackage{algorithm}
\usepackage{algorithmicx}
\usepackage{algpseudocode}
\usepackage{bm}
\usepackage{makecell}
\usepackage{color}
\usepackage{cite}
\usepackage{mathrsfs}
\usepackage{threeparttable}
\usepackage{comment}
\usepackage[mathlines,switch]{lineno}
\usepackage{lipsum}
\usepackage{booktabs}
\usepackage{bbding}
\usepackage{pifont}
\usepackage{multirow}
\usepackage{eqparbox}
\usepackage{hyperref}
\usepackage{xspace}

\newcommand{\cmark}{\ding{51}}%
\newcommand{\xmark}{\ding{55}}%
\newcommand{\bd}{\mathbf{d}}%
\newcommand{\method}{PRAM\xspace}

\begin{document}
\title{PRAM: Place Recognition Anywhere Model \\ for Efficient Visual Localization}
\author{Fei~Xue, Member, IEEE,
	Ignas~Budvytis, Member, IEEE,
	Roberto~Cipolla, Senior Member, IEEE
\thanks{Fei Xue, Ignas Budvytis, and Roberto Cipolla are with Machine Intelligence Laboratory, Department of Engineering, University of Cambridge, Trumpington Street, Cambridge CB2 1PZ, United Kingdom.\protect\\
	E-mail: \{fx221, ib255, rc10001\}@cam.ac.uk}}

\markboth{Journal of IEEE TRANSACTIONS ON PATTERN ANALYSIS AND MACHINE INTELLIGENCE, in submission}%
{Shell \MakeLowercase{\textit{et al.}}: Bare Demo of IEEEtran.cls for Computer Society Journals}

\maketitle

\begin{abstract}
	
	Visual localization is a key technique to a variety of applications, e.g., autonomous driving, AR/VR, and robotics. For these real applications, both efficiency and accuracy are important especially on edge devices with limited computing resources. However, previous frameworks, e.g., absolute pose regression (APR), scene coordinate regression (SCR), and the hierarchical method (HM), have limited either accuracy or efficiency in both indoor and outdoor environments. In this paper, we propose the place recognition anywhere model (PRAM), a new framework, to perform visual localization efficiently and accurately by recognizing 3D landmarks. Specifically, PRAM first generates landmarks directly in 3D space in a self-supervised manner. Without relying on commonly used classic semantic labels, these 3D landmarks can be defined in any place in indoor and outdoor scenes with higher generalization ability. Representing the map with 3D landmarks, PRAM discards global descriptors, repetitive local descriptors, and redundant 3D points, increasing the memory efficiency significantly. Then, sparse keypoints, rather than dense pixels, are utilized as the input tokens to a transformer-based recognition module for landmark recognition, which enables PRAM to recognize hundreds of landmarks with high time and memory efficiency. At test time, sparse keypoints and predicted landmark labels are utilized for outlier removal and landmark-wise 2D-3D matching as opposed to exhaustive 2D-2D matching, which further increases the time efficiency. A comprehensive evaluation of APRs, SCRs, HMs, and PRAM on both indoor and outdoor datasets demonstrates that PRAM outperforms ARPs and SCRs in large-scale scenes with a large margin and gives competitive accuracy to HMs but reduces over 90\% memory cost and runs 2.4 times faster, leading to a better balance between efficiency and accuracy. 
	
\end{abstract}


\section{Introduction}
\label{sec:intro}


\IEEEPARstart{V}{isual} localization aims to estimate the 6-DoF camera pose of a given image captured in a known environment. It is a fundamental task in computer vision and a key technique of various applications such as virtual/augmented reality (VR/AR), robotics, and autonomous driving. For these applications especially AR/VR, both efficiency and accuracy are important in particular on edge devices with limited computing resources. After many years of exploration, plenty of excellent methods have been proposed and can be roughly categorized as absolute pose regression (APR)~\cite{posenet,posenet-geo2017,lsg,glnet,atloc2020,mapnet,sc-wls2022,dfnet2022,nefes2024}, scene coordinate regression (SCR)~\cite{scr2013,dsac,ace2023,hscnet2020,dsac*}, and the hierarchical method (HM)~\cite{as,hfnet,lbr,as,nerfmatch}. APRs implicitly embed the map into high-level features and directly predict the 6-DoF pose with multi-layer perceptions (MLPs); they are fast, especially in large-scale scenes, but have limited accuracy due to implicit 3D information representation. Unlike APRs, SCRs regress 3D coordinates for all pixels in a query image and estimate the pose with PnP~\cite{epnp} and RANSAC~\cite{ransac1981}. SCRs work accurately and efficiently in indoor environments, but they are hard to scale up to outdoor large-scale scenes. Instead of predicting 3D coordinates directly, HMs adopt global features~\cite{netvlad,patchnetvlad2021,gem2018} to search reference images in the database and then build correspondences between keypoints extracted from query and reference images. These 2D-2D matches are lifted to 2D-3D matches and used for absolute pose estimation with PnP~\cite{epnp} and RANSAC~\cite{ransac1981} as with SCRs. HMs achieve high accuracy especially in large-scale outdoor scenes. Nevertheless, HMs have high storage cost of global and repetitive local 2D descriptors of reference images and high time cost of global reference search and exhaustive 2D-2D matching especially when graph-based matchers~\cite{superglue,imp2023,sgmnet,clustergnn} are employed. Table~\ref{tab:methods} demonstrates that none of APRs, SCRs, and HMs is both efficient and accurate in large-scale environments, prompting the question:

\textit{Can we find a both efficient and accurate solution to visual large-scale localization?}

\setlength{\tabcolsep}{4.5pt}

\begin{table}[t]
	\centering
	\caption{\textbf{Comparisons of APRs, SCRs, and HMs.}  APRs are fast but inaccurate; SCRs work well in small scenes but can hardly scale up in large-scale environments; HMs have high accuracy, but they are not efficient in terms of both time and memory.}
	\begin{tabular}{lcccc}
		\toprule
		Method &  Memory & Time & Accuracy & Large-scale\\
		\midrule 
		APRs~\cite{posenet,posenet-geo2017,glnet,mapnet,lsg,nefes2024} & \cmark & \cmark & {\color{red}\xmark} & \cmark \\
		SCRs~\cite{dsac,dsac++,localinstance,ace2023} & \cmark & \cmark & \cmark & {\color{red}\xmark} \\ 
		HMs~\cite{hfnet,lbr,as,smc} & {\color{red}\xmark} & {\color{red}\xmark}& \cmark & \cmark\\
		
		
		\bottomrule 
	\end{tabular}

\label{tab:methods}
\end{table}

\begin{figure*}[t]
	\centering
	\includegraphics[width=1.0\linewidth]{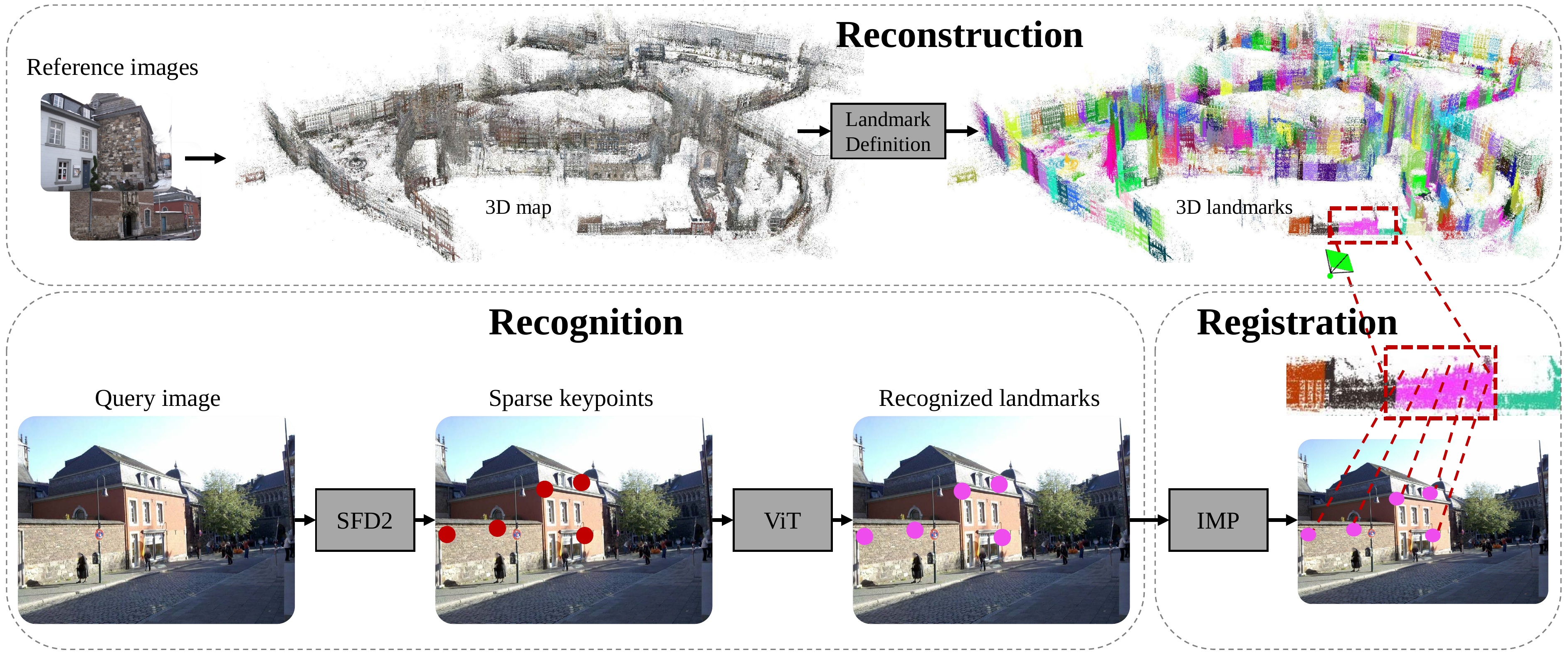}
	\caption{\textbf{Overview of the \method framework}. \method first reconstructs the 3D map of a scene from reference images and then generates landmarks in 3D space in a self-supervised manner; the recognition module utilizes sparse keypoints~\cite{sfd22023} extracted from the query image as inputs and predicts corresponding landmark labels with visual transformers; with recognized landmarks, the registration module performs landmark-wise 2D-3D matching to recover the absolute pose of the query image.}
	\label{fig:overview}
\end{figure*}

The key to efficient and accurate visual localization lies in establishing 2D-3D correspondences between the query image and the 3D map efficiently and accurately. SCRs regress 3D coordinates with high efficiency but suffer from large errors in outdoor environments. HMs adopt reference images to achieve this, resulting in low efficiency. Intuitively, objects and their spatial relationships define many landmarks associated with unique locations in 3D space. Recognizing these landmarks with a powerful network can efficiently offer us accurate 2D-3D correspondences with the guidance of landmark labels, making efficient and accurate localization possible. However, two major challenges exist. The first one is how to define landmarks in the real world. Some previous approaches define landmarks on classic objects widely used in public semantic segmentation datasets~\cite{ade20k}, e.g., building facades~\cite{lbr,localinstance}. Despite their promising performance in certain scenes, these methods can hardly be extended to other places without building facades, such as indoor environments. Additionally, many objects in the real world do not have corresponding labels in the dataset~\cite{ade20k}, resulting in the low generalization ability of these methods. The second challenge is how to recognize a large number of landmarks efficiently. For example, to recognize 512 landmarks for an image with the size of $512\times512$, a model needs to output a feature with the size of 128 Megabytes, impairing the efficiency. 

To solve the two challenges mentioned above, we propose to generate landmarks from the 3D map directly instead of classic objects. Specifically, we adopt a hierarchical clustering strategy to produce landmarks from 3D points in a self-supervised manner, allowing us to generate landmarks anywhere in indoor and outdoor scenes. Moreover, we recognize these landmarks by taking sparse keypoints as input rather than dense pixels. Sparse keypoints~\cite{superpoint,sfd22023,sift,orb,r2d2,lift} and their spatial relationships effectively reveal the structure of objects~\cite{sparc2022,sparse2023} and hence can be used to replace dense usually redundant pixels for efficient recognition. Besides, by recognizing self-defined landmarks from sparse keypoints directly, the long-term evil inconsistency problem between keyponints and semantic segmentation masks is partially mitigated (sparse keypoints are usually located at object boundaries where semantic segmentation results have large uncertainties)~\cite{smc,vso2018,lbr}. These keypoints can be taken as sparse tokens, enabling us to naturally leverage powerful transformers~\cite{attention2017} for recognition.

In this paper, we introduce the place recognition anywhere model (PRAM) for visual localization, as shown in Fig.~\ref{fig:overview}. We first extract sparse keypoints for 3D reconstruction; then we generate landmarks in 3D space automatically; finally, sparse keypoints along with their 2D coordinates are used as tokens to be fed into a transformer-based recognition module for sparse landmark recognition. At test time, sparse keypoints and predicted landmark labels are further utilized for fast landmark-wise 2D-3D matching between the query image and the map for pose estimation.

We summarize the contributions of this work as follows: We (i) first systematically analyze common frameworks for localization on public indoor and outdoor datasets, inspiring the design of PRAM, a new framework for localization via sparse landmark recognition. PRAM demonstrates that (ii) by generating landmarks in 3D space in a self-supervised manner instead of using widely-used classic object labels, \method is able to define landmarks anywhere, overcoming the limited generalization ability of classic semantic labels for localization. (iii) Representing the map with 3D landmarks rather than reference images, PRAM discards global descriptors, repetitive local descriptors, and redundant 3D points, reducing the map size significantly. (iv) Sparse recognition of landmarks allows for efficient 2D outlier keypoint removal and fast landmark-wise 2D-3D matching, leading to higher efficiency than HMs, especially in large-scale environments. The comprehensive evaluation shows that (v) PRAM generalizes well in both indoor and outdoor environments by reporting close accuracy to HMs on indoor datasets~\cite{sevenscenes2013,twelvescenes2016} and competitive performance to HMs on outdoor datasets~\cite{posenet,aachen}, but reduces over 90\% of the map size and runs 2.4 times faster. PRAM achieves a better balance between accuracy and efficiency and paves a new way towards efficient and accurate localization.

The rest of the paper is organized as follows. We review related works in Sec.~\ref{sec:related_works} and detail PRAM in Sec.~\ref{sec:method}. We evaluate the proposed method and compare it with previous approaches in Sec.~\ref{sec:experiment_setup} and Sec.~\ref{sec:experiments}. We finally conclude the paper and discuss limitations and future works in Sec.~\ref{sec:conclusion_futurework}.

\section{Related work}
\label{sec:related_works}


\textbf{Visual localization.} Visual localization methods can be roughly categorized as absolute pose regression (APR), scene coordinate regression (SCR), and the hierarchical method (HM). APRs encode images as pose features and regress the pose with an MLP. Posenet~\cite{posenet} is the first work implementing this idea. Due to its simplicity, high memory and time efficiency, a lot of variants have been proposed by introducing geometric loss~\cite{posenet-geo2017}, multi-view constraints~\cite{mapnet,lsg,glnet,pogonet2021,gtcar2022,vcr2021}, feature selection~\cite{atloc2020}, pose refinement~\cite{sc-wls2022,dfnet2022,nefes2024}, and generation of additional training data~\cite{lens2022}. However, their accuracy is still limited because of the retrieval nature of APRs~\cite{sattler2019understanding}.

Different from APRs, SCRs first regress the 3D coordinate for each pixel in the query image and then estimate the pose with PnP~\cite{epnp} and RANSAC~\cite{ransac1981}. Initially, this is achieved via random forest with RGBD data as input~\cite{scr2013}. Later, DSAC~\cite{dsac} and its variants~\cite{dsac*,dsac++} extend it to RGB input and replace random forest with powerful CNNs. More recently, sparse regression~\cite{sparse_sc2018}, hierarchical prediction~\cite{hscnet2020}, the separation of backbone and prediction head~\cite{ace2023}, to name a few, are introduced for higher accuracy and training efficiency. With explicit 2D-3D correspondences, SCRs give very accurate poses in small scenes such as indoor 7Scenes~\cite{sevenscenes2013} and 12Scenes~\cite{twelvescenes2016} datasets. However, they have limited accuracy in larger outdoor environments such as CambridgeLandmarks~\cite{posenet} and Aachen dataset~\cite{aachen} due to their large coordinate regression errors.

HMs~\cite{hfnet,lbr} conduct localization hierarchically. HMs~\cite{hfnet,lbr} first find reference images in the database, then build 2D-2D matches between keypoints extracted from query and reference images, and finally compute the pose from 2D-3D matches lifted from 2D-2D ones with PnP~\cite{epnp} and RANSAC~\cite{ransac1981}. Traditionally, bag of words (BoW) and handcrafted SIFT~\cite{sift} or ORB~\cite{orb} features are widely used for the first two steps~\cite{as}. As handcrafted features are sensitive to illumination and seasonal changes, learned local features~\cite{lift,r2d2,d2-net,disk,xfeat,superpoint,sfd22023}, e.g., SuperPoint (SP)~\cite{superpoint}, R2D2~\cite{r2d2}, SFD2~\cite{sfd22023} and global features~\cite{netvlad,gem2018,densevlad,patchnetvlad2021}, e.g., NetVLAD (NV)~\cite{netvlad}, GeM~~\cite{gem2018}, DenseVLAD~\cite{densevlad} are mainly used. To further improve the accuracy, graph-based sparse matchers~\cite{superglue,sgmnet,clustergnn,imp2023,omniglue2024}, e.g., SuperGlue (SG)~\cite{superglue}, IMP~\cite{imp2023} and dense matchers~\cite{loftr,aspanformer2022,roma2024}, e.g., LoFTR~\cite{loftr}, RoMa~\cite{roma2024} are proposed for higher matching quality. Nowadays, the combination of SP+SG+NV~\cite{superpoint,netvlad,superglue} has reported state-of-the-art accuracy on public datasets~\cite{sevenscenes2013,twelvescenes2016,posenet,aachen}. SFD2+IMP+NV~\cite{sfd22023,imp2023,netvlad} gives the close accuracy with higher efficiency. Despite the outstanding accuracy, HMs need to store a large amount of global and local descriptors to establish 2D-2D correspondences, impairing the memory efficiency. The global reference search and exhaustive 2D-2D matching further degrade the time efficiency. Some recent approaches propose to represent the map implicitly with NeRFs~\cite{nerfs2021} such as NeRFMatch~\cite{nerfmatch} and VRS-NeRF~\cite{vrsnerf2024}, despite their higher efficiency of map representation, their require additional time for rendering at test time and report relatively poor accuracy in city-scale scenes~\cite{aachen} due to the limited representation ability of NeRFs.

As with HMs, PRAM also uses sparse keypoints for 2D-3D matching. However, PRAM is essentially different from HMs. PRAM executes the fast recognition of 3D landmarks instead of performing the exhaustive global reference image search. Besides, representing the map with 3D landmarks as opposed to reference images, PRAM discards global descriptors, repetitive local descriptors, and redundant 3D points, thus has higher memory efficiency. Moreover, PRAM establishes 2D-3D matches with the guidance of predicted landmarks as opposed to exhaustive 2D-2D matching, resulting in further improvements in time efficiency.


\textbf{Visual semantic localization.} Compared with pixels, semantics are more robust to appearance changes, therefore researchers have tried almost all efforts to apply semantics to localization. Most of these methods~\cite{smc,vls,ssm,lln,fgsn,slam++2013} utilize explicit semantic labels predicted by segmentation networks~\cite{segnet2017,deeplab3plus} to filter unstable keypoints and semantic-inconsistency matches. However, improvements are limited due to the following reasons: (i) Sparse keypoints detected from object boundaries usually have large segmentation uncertainties. (ii) Segmentation results of images under challenging conditions (e.g., low-illumination) contain large errors~\cite{sfd22023}. (iii) The manually defined strategy of filtering keypoints according to object classes has low generalization ability. For example, trees are useful for short-term localization~\cite{orbslam2,beyondtracking2019,savo2019,dso2017} but useless for long-term localization~\cite{lbr,smc,as}; trees can be used for coarse localization but may cause fine localization errors.

More recently, some of the aforementioned limitations have been mitigated. SFD2~\cite{sfd22023} embeds semantics into the detector and descriptor implicitly and extracts semantic-aware features directly in an end-to-end fashion. The implicit embedding of semantics avoids semantic uncertainties caused by the usage of explicit labels at test time. However, these classic semantic labels can hardly be used for global place recognition. Instead, some works use building facades as global instances for coordinate regression~\cite{localinstance,budvytis2019semantic} and fast reference image search~\cite{lbr}. These global instances are defined only on building facades, so they can hardly be applied to places without building facades such as indoor scenes. In short, although semantics have great potential to improve localization performance, segmentation errors, high sementation uncertainties at object boundaries, and low generalization ability in indoor and outdoor scenes degrade the gains of using semantics. 

Different from the abovementioned approaches, PRAM generates landmarks from 3D points directly, so PRAM can work in any place. In addition, landmark labels are defined on 3D points as opposed to objects, so the problem of uncertainties from object boundaries is alleviated. Besides, these generated landmarks have both global and local properties and thus can be used for both global place recognition and local matching. Moreover, whether a 2D keypoint is an inlier or not is determined by whether it has a corresponding 3D point in the map or not. This can be achieved by the recognition module directly without relying on any manually defined strategy which usually has low generalization ability.

\begin{figure*}[t!]
	\centering
	\includegraphics[width=1.\linewidth]{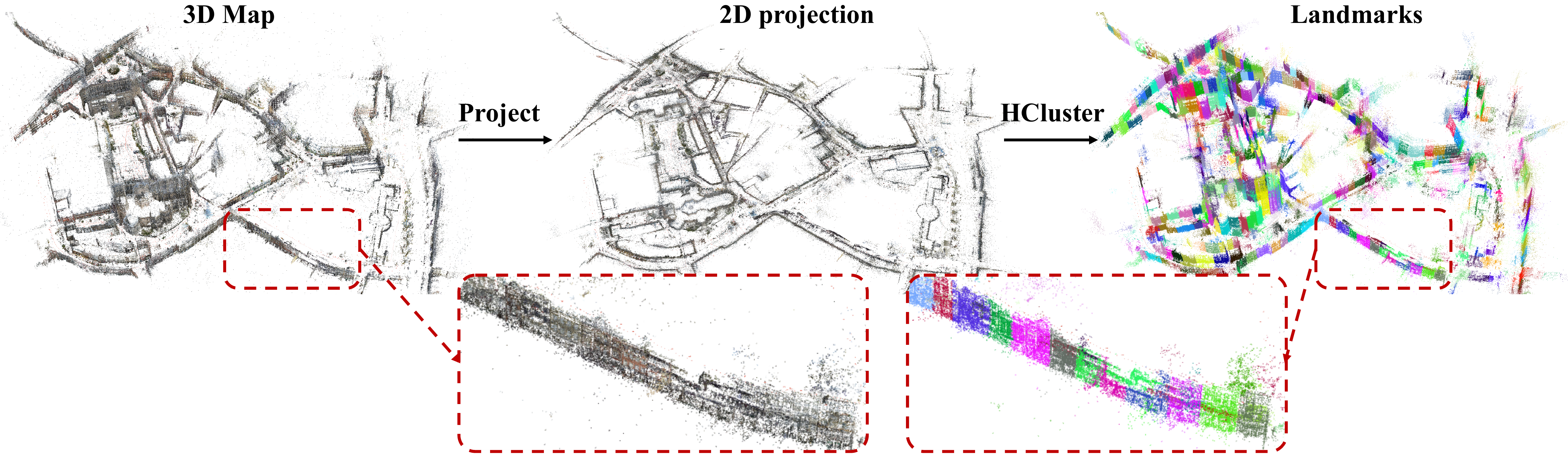}
	\caption{\textbf{Landmark generation on the Aachen dataset~\cite{aachen}}. 3D points in the map are first projected to the ground plane as 2D projections. Then we employ a hierarchical approach to perform clustering on 2D projections based on spatial connections. Compared with reference images, the landmark map represents the large-scale Aachen city~\cite{aachen} in a compact way with 512 landmarks (best view in color).}
	\label{fig:landmark_definition}
\end{figure*}

\section{Method}
\label{sec:method}

In this section, we first describe the 3D landmark generation strategy and the structure of the 3D map represented by landmarks in Sec.~\ref{sec:method:landmark_definition} and Sec.~\ref{sec:method:representation}, respectively. Then we introduce the progress of recognizing landmarks with sparse keypoints in Sec.~\ref{sec:method:sparse_rec} and the usage of predicted landmarks for localization in Sec.~\ref{sec:method:localization}.

\subsection{3D landmark generation}
\label{sec:method:landmark_definition}

\textbf{3D reconstruction.} We first reconstruct a 3D map represented by $m$ 3D points $\mathcal{X}=\{X_1,...,X_m\}$ ($X_i \in R^{3}$) from images in the database with current state-of-the-art 3D reconstruction library Colmap~\cite{colmap2016}. Instead of using handcrafted features~\cite{sift,orb} and nearest matching, we adopt deep local feature SFD2~\cite{sfd22023} and graph-based matcher IMP~\cite{imp2023} to extract sparse keypoints and build correspondences, respectively. SFD2~\cite{sfd22023} embeds semantics into features implicitly and is more robust appearance changes for long-term localization; IMP~\cite{imp2023} gives better performance and runs faster than prior matchers~\cite{superglue,sgmnet,clustergnn}. Note that other deep features~\cite{superpoint,d2-net,r2d2,disk} and matchers~\cite{superglue,sgmnet,clustergnn} can also be used in our framework. Each image $I\in R^{3 \times H\times W}$ ($H$ and $W$ are the height and width) is first fed into SFD2 network to extract sparse keypoints with each represented as $p=(u,v,\bd^{2D})$ where $(u,v)$ is the 2D coordinate and $\bd^{2D}\in R^{128}$ is the 2D descriptor. We employ the default setup of SFD2 to retain no more than 4k keypoints with scores over $0.005$ for each image in both the reconstruction and localization processes. 

Due to insufficient observations and wrong matches, outlier 3D points exist, impairing the map quality. To handle this, we only keep 3D points which are spatially consistent with their neighbors, as $\mathcal{X}=\{X | V(X, \lambda_n) \leq \lambda_v\}$ where $V(X, \lambda_n)$ computes the covariance of a 3D point $X$ with its $\lambda_n$ nearest neighbors and $\lambda_v$ is the threshold. $\lambda_n$ and $\lambda_v$ are set to 20 and 0.2 in our experiments. 


\textbf{Landmark generation.} As shown in Fig.~\ref{fig:landmark_definition}, we define landmarks on 3D points directly in a self-supervised manner. More precisely, we perform hierarchical clustering by merging 3D points from bottom to up according to their spatial distances. As most objects in the real world such as bookshelves, tables, trees, and building facades, are vertical to the ground, we project each 3D point $X$ to the 2D ground plane and then execute clustering on their 2D projections, allowing us to better maintain the completeness of objects, especially in outdoor environments. This process can be expressed as: 
\begin{align}
	X^{2D}_i&=\begin{bmatrix}
		1 & 0 & 0 \\
		0 & 1 & 0
	\end{bmatrix}X_i, \\
	 \{L_1,...,L_{\lambda_l}\}&= f_{hcluster}(\{X^{2D}_1,...,X^{2D}_k\}, \lambda_l).
\end{align}
$f_{hcluster}(\cdot)$ is the function performing hierarchical clustering on projected 2D point set $\{X^{2D}_1,...,X^{2D}_k\}$ and assigns each 3D point $X_i$ a landmark label $L_i$. As with BIRCH~\cite{birch1996}, we employ the same metric defined on the clustering feature to decide if a new point should be merged into an existing leaf or taken as a new leaf. We refer to BIRCH~\cite{birch1996} for more details about the clustering process. The merging process stops until reaching the given number of landmarks $\lambda_l$ which is determined by the size of the scene. Compared with other self-supervised methods such as k-means++~\cite{kmeans++2007}, $f_{hcluster}(\cdot)$ iteratively reduces and clusters through hierarchy clustering feature tree and works better for large-scale data with low dimensions. Moreover, some objects are still split into different landmarks even if with the spatial constraints. The hierarchical manner in $f_{hcluster}(\cdot)$ is flexible to accept other information such as object consistency provided by SAM~\cite{sam2023} to mitigate the problem. Currently, we find results of SAM are noisy especially for outdoor scenes, leading to inconsistent masks of the same objects between multi-view images. In the future, we will reduce the noise of SAM results and add object-level constraints to $f_{hcluster}(\cdot)$.
\begin{figure}[t]
	\centering
	\includegraphics[width=1.\linewidth]{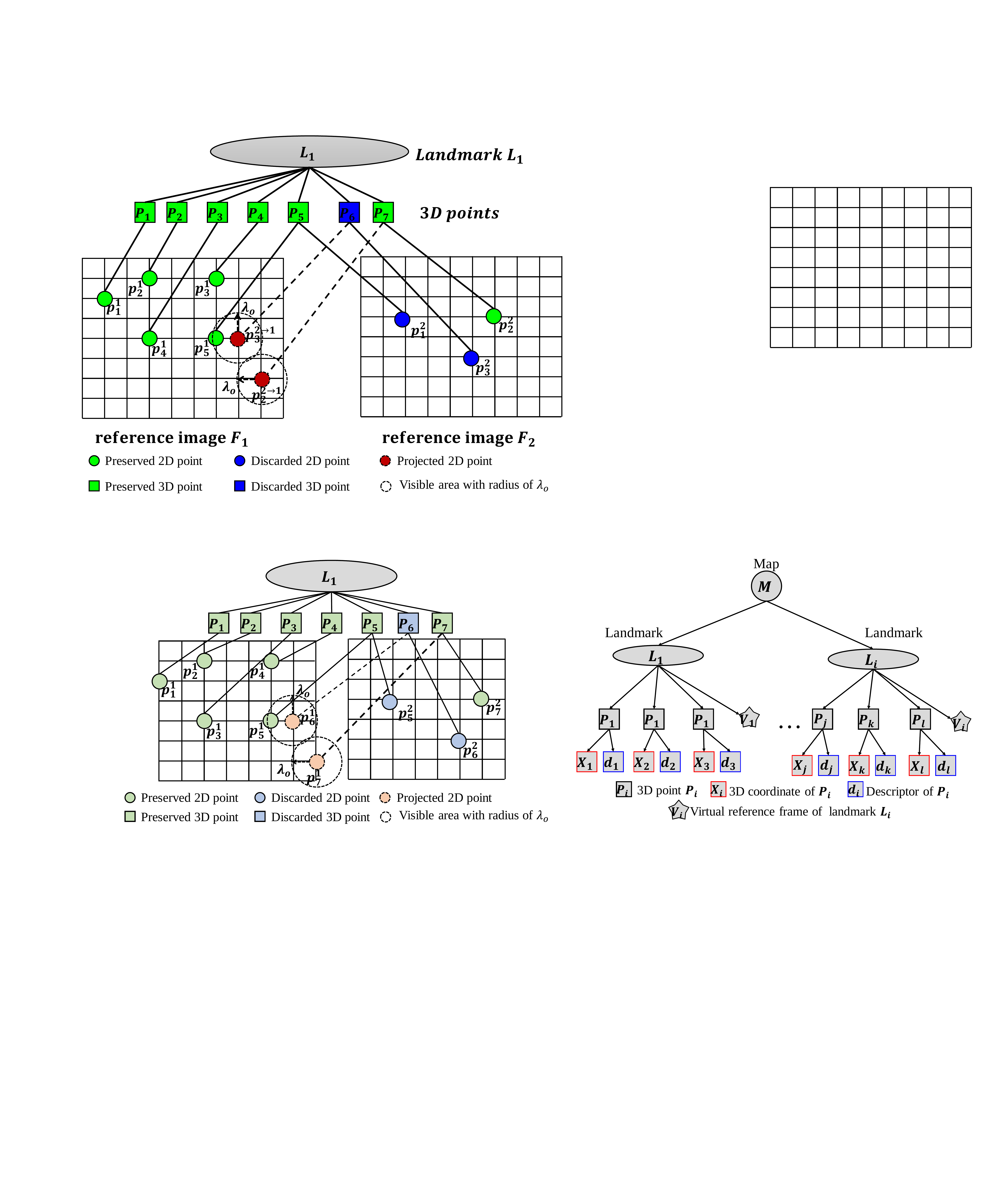}
	\caption{\textbf{The structure of map represented by 3D landmarks}. A 3D map $\mathcal{M}$ is represented by a number of landmarks $\mathcal{L} = \{L_1,...,L_{\lambda_l}\}$ ($\lambda_l$ is the total number of landmarks). Each landmark $L_i$ contains several 3D points $\mathcal{P}_i=\{P_1,...,P_k\}$ and a virtual reference frame $V_i$. Each 3D point $P_i$ consists of its 3D coordinate $X\in R^{3}$ and descriptor $\bd\in R^{128}$.}
	\label{fig:map_structure}
\end{figure}

\subsection{Map representation}
\label{sec:method:representation}

With reconstructed 3D points $\{X\}$ and generated landmarks $\{L\}$, we name the map as $\mathcal{M}$ and reorganize its structure to fit the PRAM localization framework.

\textbf{Structure of Map.} As shown in Fig.~\ref{fig:map_structure}, the map $\mathcal{M}$ is represented by $\lambda_l$ landmarks $\{L_1,...,L_{\lambda_l}\}$; each landmark $L_i$ contains some 3D points $\mathcal{P}_i=\{P_1,...,P_j\}$ and a virtual reference frame (VRF) $V_i$; each 3D point $P_i$ is represented by its 3D coordinates $X_i$ and 3D descriptor $\bd_i\in R^{128}$. 

\textbf{3D Descriptor.} Each 3D point $P_i$ is assigned with a 3D descriptor to build 2D-3D matches between query images and the map for registration. In order to mitigate the domain differences between 2D and 3D points, we select 3D descriptors carefully for each 3D point $P_i$ from its 2D observations in the reconstruction process, as:
\begin{align}
	\bd_i = \operatorname*{argmin}_{\bd} f_{md}(\bd, \bd^{2D}_1,..., \bd^{2D}_j),\bd \in \{\bd^{2D}_1,...,\bd^{2D}_j\}.
\end{align}
For point $P_i$, its descriptor $\bd_i$ is the 2D descriptor which has the smallest median distance computed by $f_{md}(\cdot)$ to all other 2D descriptors $\{\bd^{2D}_1,...,\bd^{2D}_j\}$ of keypoints observing $P_i$ on reference images. The median distance has higher statistical robustness and has been proved effective to appearance changes~\cite{orbslam2} especially when query and reference images are captured at different times. 

\textbf{Virtual reference frame (VRF).} We assign each landmark $L_i$ a virtual reference frame (VRF) $V_i$ for two reasons. First, the pose of a VRF $V_i$ can be taken as an approximation of the location of the corresponding landmark $L_i$ in the 3D space. In the localization process, such an approximation can be used as a coarse location of the query image. Second, VRFs allow for the projection of 3D points onto an image plane in the matching process,  mitigating the domain gap between 2D and 3D points. Note that VRFs are different from reference images used in HMs~\cite{lbr,hfnet} as VRFs do not contain any global or local descriptors. The VRF $V_i$ assigned for landmark $L_i$ should satisfy the requirement that $V_i$ observes the major part of 3D points belonging to $L_i$. Therefore, we choose $V_i$ as:
\begin{align}
	\label{eq:vrf}
	V_i = \operatorname*{argmax}_{F} f_{obs}(F, L_i).
\end{align}
$V_i$ is the reference image $F$ which observes the largest number of 3D points  belonging to landmark $L_i$. $f_{obs}(F, L_i) = \frac{|\{X_i | X_i\in\mathcal{X}_F\cap\mathcal{X}_i\}|}{|\{X_i | X_i\in\mathcal{X}_i\}|}$ computes the ratio of the number of observed points $|\{X_i | X_i\in\mathcal{X}_F\cap\mathcal{X}_i\}|$ and the total number of points $|\{X_i | X_i\in\mathcal{X}_i\}|$ in $L_i$. Each VRF $V_i$ has the intrinsic ${K_i}$ and extrinsic parameters $T_i$. The virtual reference frames are used in the localization process as discussed in Sec.~\ref{sec:method:localization}.

\begin{figure}[t]
	\centering
	\includegraphics[width=1.\linewidth]{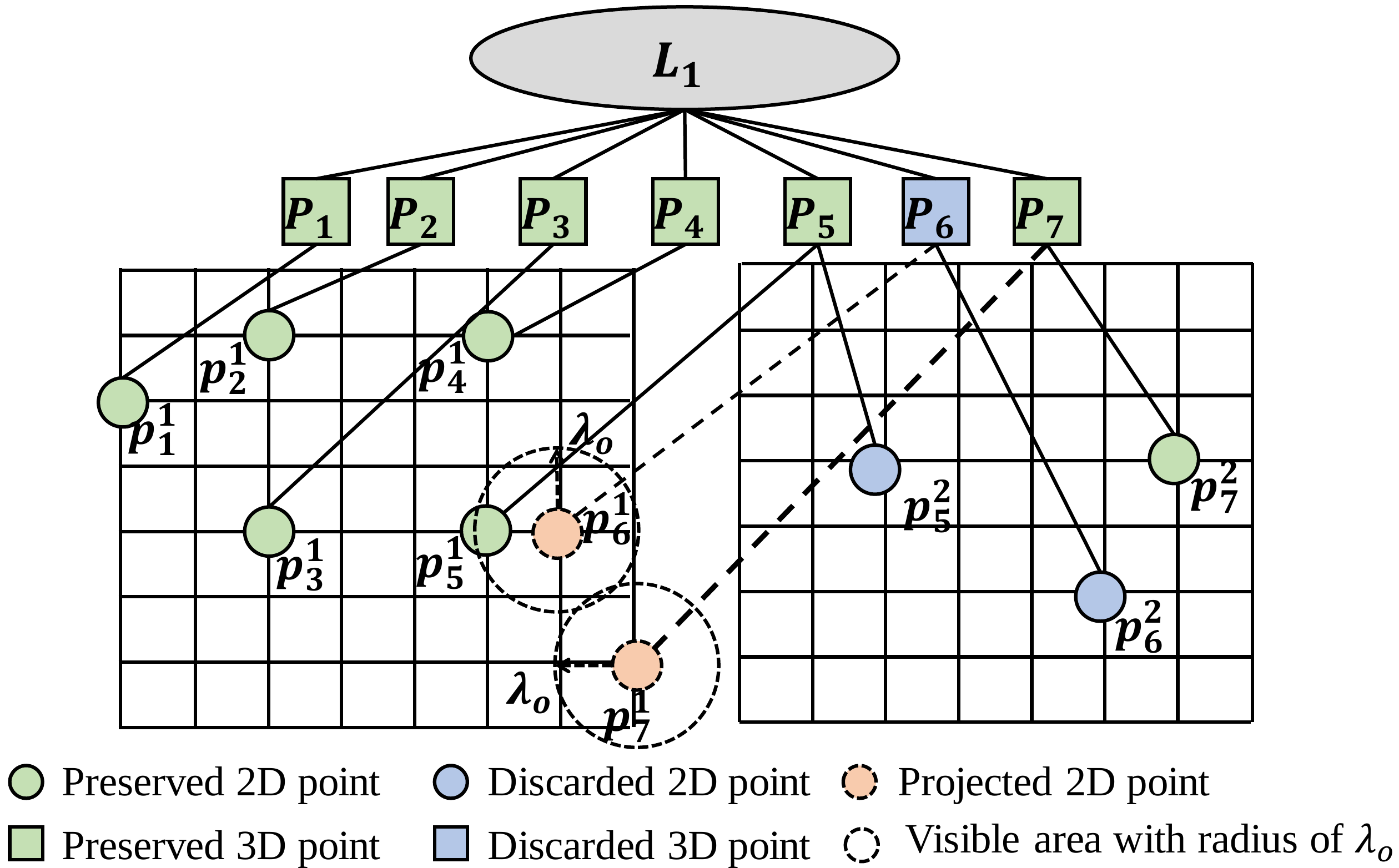}
	\caption{\textbf{Adaptive landmark-wise 3D point pruning}. Taking landmark $L_1$ with 3D points of $\{P_1,...,P_7\}$ and two reference images $F_1, F_2$ for example, we first retain points $P_{1:5}$ as they can be observed by $F_1$. As $P_5$ is already observed by $F_1$ as $p^{1}_5$, its observation on $F_2$ as $p^{2}_5$ is removed. The 2D projection $p^{1}_6$ on frame $F_1$ of $P_6$ has a keypoint $p^{1}_5$ within a circle visible area with radius of $\lambda_o$, so $P_6$ is removed from landmark $L_1$ and its 2D observation $p^{2}_6$ is also removed from $F_2$. The 2D projection $p^{1}_7$ on frame $F_1$ of $P_7$ has no candidates within circle of radius $\lambda_o$, so $P_7$ and $p^{2}_7$ are retained.}
	\label{fig:map_sparsification}
\end{figure}

\begin{figure*}[htbp]
	\centering
	\includegraphics[width=1.\linewidth]{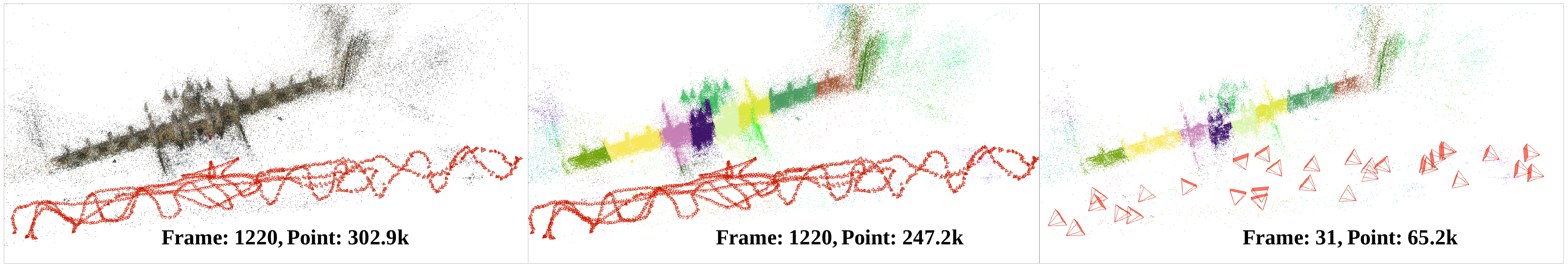}
	\caption{\textbf{The original map and the map represented by 3D landmarks}. The left shows 3D points and reference images of the map reconstructed with Colmap~\cite{colmap2016} of Kings College in CambridgeLandmarks~\cite{posenet}. The middle visualizes the 3D points after removing locally inconsistent ones and generated 3D landmarks. The right presents the virtual reference frames (VRFs) and 3D points after adaptive landmark-wise 3D pruning. The number of both images and 3D points are also included.}
	\label{fig:sparse_map_kingscollege}
\end{figure*}

\begin{figure*}[t]
	\centering
	\includegraphics[width=1.\linewidth]{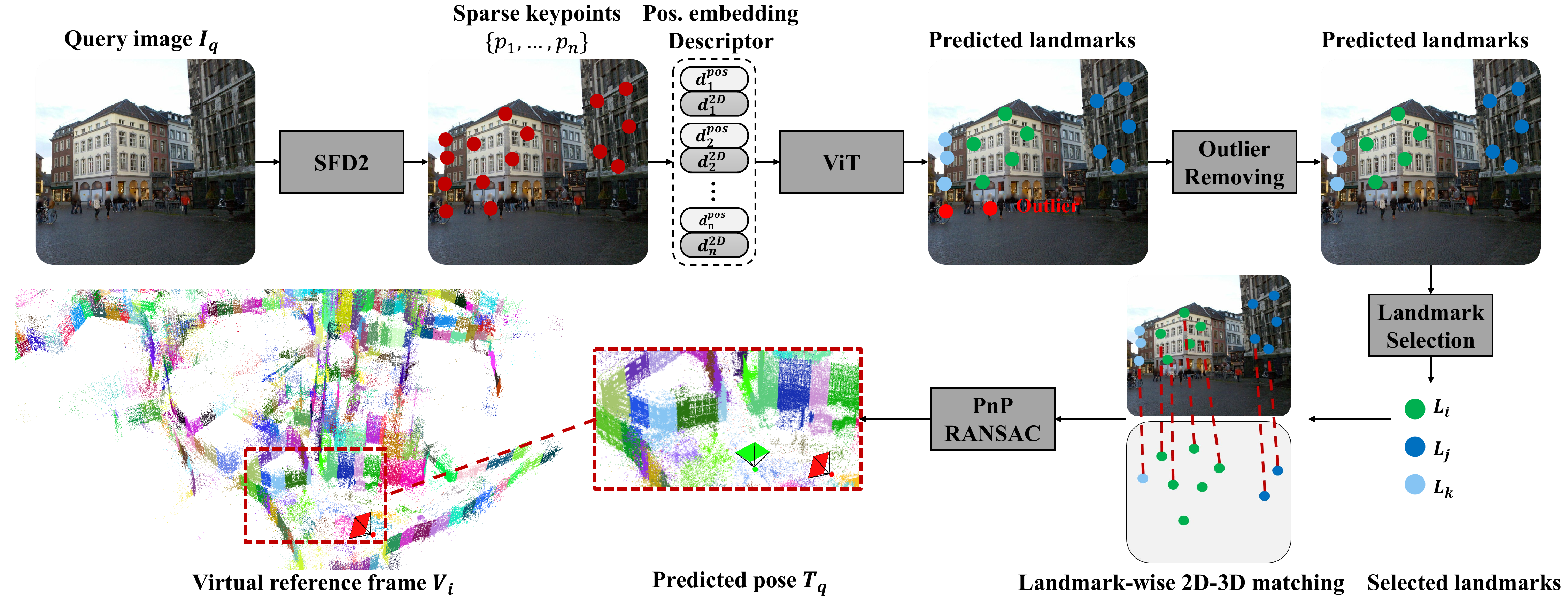}
	\caption{\textbf{Pipeline of localization by sparse recognition}. For a query image, \method first extracts sparse SFD2 keypoints~\cite{sfd22023}. Then, descriptors and position embeddings of these keypoints are fed into a visual transformer for sparse landmark prediction. Next, outliers with predicted labels of 0 are discarded for efficient registration. The remaining keypoints and their landmark labels are used to choose candidate landmarks for registration. For each candidate landmark, landmark-wise matching is employed to build 2D-3D correspondences between the query and virtual reference frame. Finally, PnP~\cite{epnp} and RANSAC~\cite{ransac1981} are used to estimate the absolute pose. Poses of the query image (\textcolor{green}{green}) and the virtual reference frame (\textcolor{red}{red}) are visualized.}
	\label{fig:pipeline}
\end{figure*}

\textbf{Adaptive landmark-wise 3D point pruning.} Landmark labels provide strong priors for 2D-3D point matching between query images and the map, so each landmark does not need to preserve a large number of usually redundant 3D points. Therefore, we adaptively remove these redundant 3D points for each landmark independently.

Initially, let $F_1,...,F_n$ and $\mathcal{X}_{F_1},...,\mathcal{X}_{F_n}$ be reference images and corresponding observed 3D points in the landmark $L_i$. $F_1, ..., F_n$ are sorted in descending order according to their numbers of observations ($|\mathcal{X}_{F_1}| \geq |\mathcal{X}_{F_2}| \geq ... \geq |\mathcal{X}_{F_n}|$). Note that according to Eq~(\ref{eq:vrf}), intrinsic and extrinsic parameters of $F_1$ are used to construct the VRF of $L_i$. For the first iteration, we choose $F_1$ as the reference image and initialize the preserved keypoints for $F_1$ as $\mathcal{X}^p_{F_1}=\mathcal{X}_{F_1}$. Next, we remove repetitive points on $F_2$ which can be observed by $F_1$ as:
\begin{align}
	\mathcal{X}^{p}_{F_2}=\mathcal{X}_{F_2} -  f_{overlap}(F_1, F_2, \mathcal{X}_{F_1}, \mathcal{X}_{F_2}, \lambda_o).
\end{align}
Points on $F_2$ with reprojection distances to any point on $F_1$ smaller than $\lambda_o$ are discarded, as shown in Fig.~\ref{fig:map_sparsification}. This strategy effectively removes redundant points with high overlap with slight localization accuracy loss (as discussed in Sec.~\ref{sec:experiments}). $\mathcal{X}^{p}_{F_2}$ is the set of remaining points observed by $F_2$ and $f_{overlap}(\cdot)$ computes the overlapped points within the distance threshold of $\lambda_o$. For the i$th$ iteration, the updating process is as:
\begin{align}
	\mathcal{X}^{p}_{F_{i}}= \mathcal{X}_{F_{i}} - f_{overlap}(F_{1:i-1}, F_{i}, \mathcal{X}^{p}_{F_1:F_{i-1}}, \mathcal{X}_{F_{i}}, \lambda_o).
\end{align}
 %
Finally, we collect preserved points from all processed images as $\mathcal{X}^{p}_{F_1}\cup...\cup\mathcal{X}^{p}_{F_n}$ to replace the original full point set $\mathcal{X}_{F_1}\cup...\cup\mathcal{X}_{F_i}$ as the 3D points in landmark $L_i$. Compared with previous methods solving a K-cover~\cite{kcover2010} or quadratic programming~\cite{qp2013} problem for the whole map, this strategy can simply be applied to each landmark independently.

Fig.~\ref{fig:sparse_map_kingscollege} shows the original 3D map consisting of 3D points and reference images (left), generated landmarks from 3D points (middle), and the map represented by 3D landmarks and virtual reference frames (right). As reference images used in HMs~\cite{hfnet} do not contain the scale information of the scene, HMs usually have a large number of redundant reference images and 3D points. In contrast, landmarks defined directly in 3D space preserve the absolute scale of the scene. By assigning each landmark a virtual reference frame and performing the adaptive landmark-wise 3D point running, \method effectively discards redundant reference images (31 vs. 1,220) and 3D points (65.2k vs. 302.9k).

\subsection{Sparse recognition}
\label{sec:method:sparse_rec}

After each 3D point $P$ is associated with a landmark label $L$, all 2D keypoints observing $P$ are automatically assigned with the landmark label $L$. 2D keypoints without 3D correspondences due to non-successful triangulation or outlier filtering are assigned with label 0, indicating that they have no corresponding 3D points in the map, also known as outliers. This strategy allows \method to leverage all potential informative keypoints including those from trees for recognition but remove them easily for localization at test time.


In the training process, to better utilize the spatial connections of extracted keypoints, 2D coordinate $(u,v)$ of each keypoint $p$ is encoded as a high level position feature vector with position encoder $f_{pos}(\cdot)$. The position feature vector and original descriptor $\bd^{2D}$ are fed into a visual transformer $f_{ViT}(\cdot)$ for sparse landmark recognition. The whole process can be represented as:
\begin{align}
	\bd^{'}_i &= \bd^{2D}_i + f_{pos}(u_i, v_i), \\
	S &= f_{ViT}(\bd^{'}_1, \bd^{'}_2,...,\bd^{'}_m), \\
	Loss &= - \frac{1}{m}\sum_{i=1}^{m} w(x_i)\sum_{j=1}^{\lambda_l + 1}S^{gt}_{ij}log(S_{ij}),
\end{align}
$S\in R^{m\times(\lambda_l+1)}$ and $S^{gt}\in R^{m\times(\lambda_l+1)}$ are predicted and ground-truth recognition confidences. $S_{ij}$ indicates the confidence of $p_i$ belonging to landmark $L_j$ after \textit{softmax}. $\lambda_l$ is the total number of landmarks excluding 0. We adopt a weighted CrossEntropy loss with the weight of $w_i$ for each keypoint $p_i$ to balance the inlier and outlier keypoints. $w_i$ is set to $\frac{m_0}{m}$ if the ground-truth landmark label of $x_i$ is not $0$ otherwise $1 - \frac{m_0}{m}$ ($m_0$ is the number of keypoints with label 0 and $m$ is the total number of keypoints). 



\begin{algorithm}[t]
	\caption{Localization by sparse landmark recognition}
	\begin{algorithmic}[1]
        \State \textbf{Input: the query image $I_q$ and 3D map $\mathcal{M}$ consisting of landmarks $\mathcal{L}$, VRFs $\mathcal{V}$, and 3D points $\mathcal{P}$}

        \State \textbf{Step 1: extract SFD2 keypoints and predict landmarks}
        \State $\mathcal{P}_q = SFD2(I_q)$ 
        \State $S_q = f_{ViT}(\mathcal{P}_q)$
        \State $L_q = f_{sel}(S_q)$ \Comment{Compute landmarks}
        \State \textbf{Step 2: remove outliers with $f_{or}$ and sort landmarks}
        \State $\mathcal{P}^{'}_q, \mathcal{L}^{'}_q, S^{'}_q = f_{outlierremove}(\mathcal{P}_q, L_q, S_q, \lambda_s)$
        \State $\mathcal{P}^{''}_q \mathcal{L}^{''}_q, S^{''}_q = sort(\mathcal{P}^{'}_q, L^{'}_q, S^{'}_q)$
        
		\State \textbf{Step 3: perform progressive verification with sorted landmarks $\mathcal{L}^{''}$} \textbf{and corresponding 2D keypoints $\mathcal{P}^{''}$}
		
		\For{$l$ in $\mathcal{L}^{''}$}
		
		\State $V_{l}, \mathcal{P}_{l} = f_{retrieve}(\mathcal{M}, l)$  \Comment{Retrieve VRF and 3D points of landmark $l$ from 3D map $\mathcal{M}$}
		\State $\mathcal{P}^{2D}_{l} = f_{proj}(\mathcal{P}_{l}, V_{l})$ \Comment{Project 3D keypoints onto VRF }
		\State $M_l=IMP(\mathcal{P}^{''}_{ql}, \mathcal{P}^{2D}_{l})$ \Comment{Build matches with IMP~\cite{imp2023}}
		\State $T_l, n_{l}=f_{epnp+ransac}(\mathcal{P}^{''}_{ql}, \mathcal{P}_{l}, M_l)$ \Comment{Estimate the pose $T_l$ and compute the number of inliers $n_l$ with EPnP+RANSAC}
		\If{$n_l\geq \lambda_i$} \Comment{A pose with sufficient inliers}
		\State $T_{init} \leftarrow T_l, V_{init} \leftarrow V_l$ \Comment{Return the pose and VRF}
		\State \textbf{break}
		\EndIf
		\EndFor
		
		\State \textbf{Step 4 (optional): refine the initial pose $T_{init}$}
		\State $\mathcal{X}_{refine} = f_{covis}(T_{init}, V_{init}, \mathcal{M})$ \Comment{find more 3D points with covisibility for refinement}
		\State $M_{refine} = f_{geo}(\mathcal{P}^{''}_q, \mathcal{X}_{refine}, T_{init})$ \Comment{Find more 2D-3D matches by projecting covisible 3D points $\mathcal{X}_{refine}$ onto the query image with the initially estimated pose $T_{init}$}
		\State $T_q = f_{refine}(\mathcal{P}^{''}_q, \mathcal{X}_{refine}, \mathcal{M}_{refine})$ \Comment{Refine the pose with more 2D-3D matches}
		
	\end{algorithmic}
	\label{alg:loc_by_rec}
\end{algorithm}

\subsection{Localization by recognition}
\label{sec:method:localization}

As shown in Fig.~\ref{fig:pipeline}, in the localization process, given a query image $I_q$, we first extract sparse SFD2 keypoints~\cite{sfd22023} denoted as $\mathcal{P}_q$, then predict corresponding landmark labels $\mathcal{L}_q$ and confidences $\mathcal{S}_q$ for all keypoints with $f_{ViT}(\cdot)$. The extracted keypoints $\mathcal{P}_q$, predicted landmark labels $\mathcal{L}_q$, and confidences $\mathcal{S}_q$ are used for localization. 

\textbf{Outlier removal.} Extracted sparse keypoints usually contain many keypoints without 3D correspondences in the map. These keypoints might be useful for recognition, but are useless for registration. Fortunately, in the training process, these keypoints are assigned with label 0, so in the localization process, any keypoint $p_i\in\mathcal{P}_{q}$ with confidence of $S_{qi}$ being label 0 over the threshold of $\lambda_s$ can be easily identified as an outlier and removed. $\lambda_s$ is set to 0.9 in our experiments. We adopt this universal metric that if a keypoint has a corresponding 3D point in the map for outlier removal. Compared with previous widely-used strategies of utilizing object labels as priors (e.g., keypoints from buildings are inliers and keypoints from trees are not)~\cite{lbr,fgsn,smc,svl2017}, our method is simpler and generalizes well in any environments. The left keypoints along with their landmarks and confidences are denoted as $\mathcal{P}^{'}_q$, $\mathcal{L}^{'}_q$ and $\mathcal{S}^{'}_q$, respectively. 

\textbf{Landmark selection.} A query image usually observes several landmarks in the map, as shown in Fig.~\ref{fig:pipeline}, yet not all of them are involved in registration. To accelerate the verification process, we sort recognized landmarks in $\mathcal{L}^{'}_q$ in descending order according to their mean confidence $\mathcal{S}^{'}_q$ of belonging to a certain landmark. These sorted landmarks, corresponding confidences, and keypoints are denoted as $\mathcal{L}^{''}_q$, $\mathcal{S}^{''}_q$, and $\mathcal{P}^{''}_q$, respectively.

\textbf{Landmark-wise 2D-3D matching.} For each candidate landmark label $l\in\mathcal{L}^{''}_q$, its 2D keypoints $\mathcal{P}^{''}_{ql}\subseteq\mathcal{P}^{''}_{q}$ and 3D points $\mathcal{P}_{l}\subseteq\mathcal{P}$ belonging to label $l$ are used to build 2D-3D matches. We adopt the graph-based matcher IMP~\cite{imp2023} to find 2D-3D matches. However, due to domain differences between 2D and 3D points, direct 2D-3D matching is not robust. Instead, we reproject 3D points $\mathcal{P}_{l}$ in landmark $l$ on its virtual reference frame $V_l$ with corresponding intrinsic $K_l$ and extrinsic parameters $T_l$ as $\mathcal{P}^{2D}_{l}$. Keypoints from query images and the projected 2D keypoints from 3D map are used as inputs to IMP to produce 2D-3D matches. In this stage, we progressively verify each landmark $l$ in $\mathcal{L}^{''}_q$ to avoid exhaustive matching and reduce the time cost. 

\begin{figure*}[t]
	\centering
	\includegraphics[width=1.\linewidth]{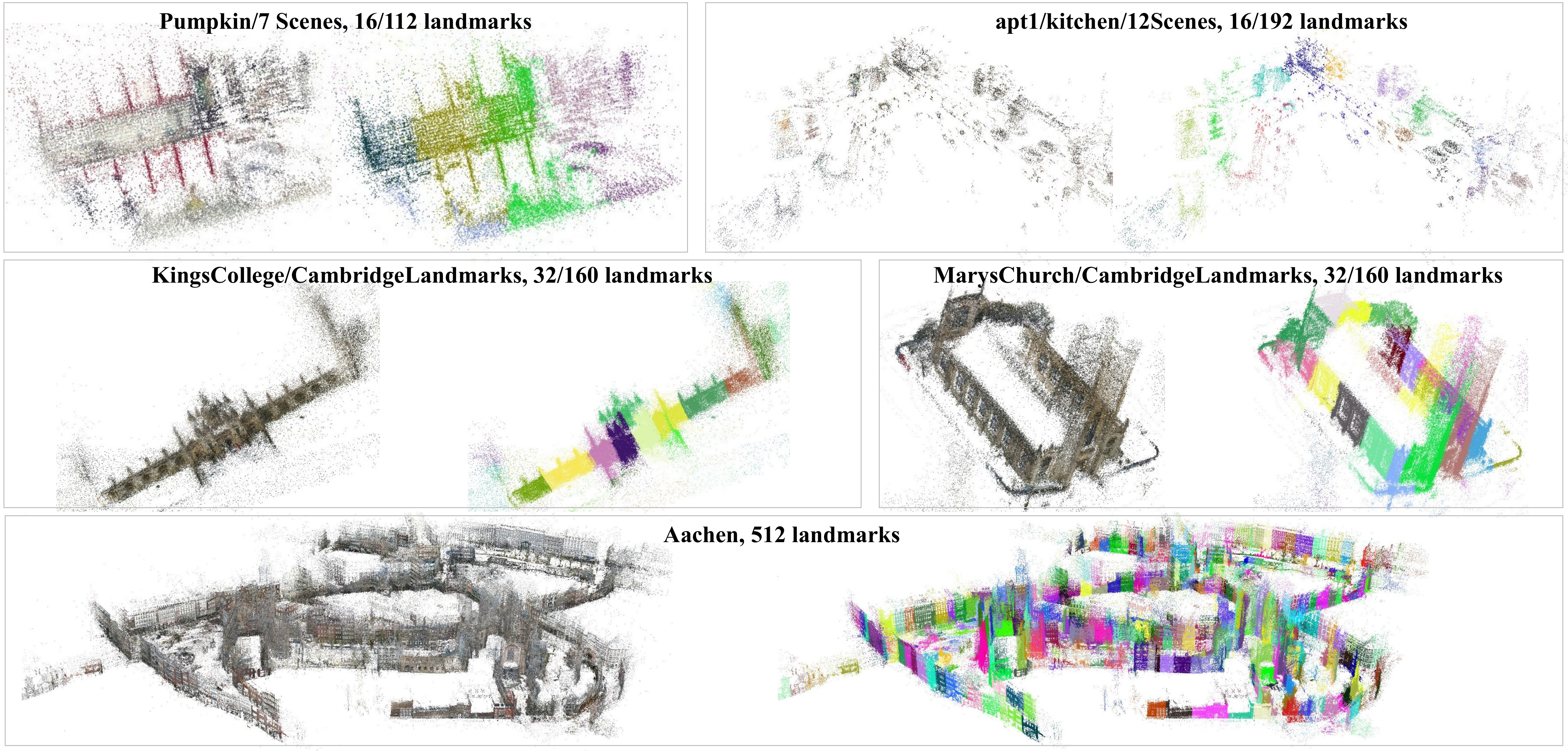}
	\caption[3D map and landmarks]{\textbf{3D point map and landmarks.} We visualize 3D points (left) and landmarks (right) in indoor (pumpkin in 7Scenes~\cite{sevenscenes2013}, apt1/kitchen in 12Scenes~\cite{twelvescenes2016}), outdoor (Kings College and Marys Church in CambridgeLandmarks~\cite{posenet}) and city-scale scenes (Aachen~\cite{aachen}). We take each dataset as a whole, so the visualized landmarks are only from example scenes. The number of landmarks in example scenes and the total number of landmarks in the dataset are also included.}
	\label{fig:map_3d_seg}
\end{figure*}

\textbf{Registration.} The 2D-3D matches are fed into EPnP~\cite{epnp} + RANSAC~\cite{ransac1981} to estimate the initial absolute pose $T_{init}\in R^{3\times4}$ of the query image. If the number of inliers is over $\lambda_i$, the verification is deemed successful, otherwise, the next candidate landmark in $\mathcal{L}^{''}_q$ is used until a successful pose is found or all candidate landmarks are explored. The initially estimated pose $T_{init}$ from a single landmark may not be very precise due to insufficient matches, so $T_{init}$ is further used to find more co-visible 2D-3D matches to obtain the finally refined pose $T_q$ as in~\cite{lbr}. The whole process is described in Algorithm~\ref{alg:loc_by_rec}.

\section{Experiment Setup}
\label{sec:experiment_setup}

In this section, we give details of the implementation, datasets, metrics, and baselines used for evaluation.

\textbf{Implementation.} We implement the recognition model $f_{ViT}$ with 15 ViT blocks consisting of multi-head self-attention layers on PyTorch~\cite{pytorch}. The number of heads and the dimension of the hidden feature are set to 4 and 256, respectively. $f_{pos}$ is a 4-layer MLP with the hidden dimension of 32, 64, 128, and 256. All models are trained with Adam optimizer~\cite{adam} with a batch size of 32 on the NVIDIA RTX 3090 GPU. Hyper-parameter $\lambda_o$ is set to 25; $\lambda_l$ is set to 112, 192, 160, 512, $\lambda_c$ is set to 20, 20, 20, 50, and $\lambda_i$ is set to 64, 64, 128, 128 for 7Scenes~\cite{sevenscenes2013}, 12Scenes~\cite{twelvescenes2016}, CambridgeLandmarks~\cite{posenet}, and Aachen~\cite{aachen,visbenchmark} datasets. The source code is publically available at \url{https://github.com/feixue94/pram}.

\textbf{Dataset.} \method is evaluated on a variety of widely-used public indoor and outdoor datasets including 7Scenes~\cite{sevenscenes2013}, 12Scenes~\cite{twelvescenes2016}, CambridgeLandmarks~\cite{posenet}, and Aachen~\cite{aachen} datasets. 7Scenes and 12Scenes are indoor scenes with 7 and 12 individual rooms, respectively. Each room has a size of around $2m \times 2m \times 2m$. CambridgeLandmarks dataset consists of 5 individual scenes captured at the center of Cambridge city. Each has an area size of about $20m \times 50m$ (for simplicity, we omit the height for outdoor scenes). The Aachen dataset contains images captured at different seasons with various viewpoints in the Aachen city. It has an area size of about $3km \times 2km$. Dynamic objects, illumination and season variations between query and reference images make the localization very challenging. Different from previous APRs~\cite{posenet,posenet-geo2017,lsg,glnet,sc-wls2022,atloc2020} and SCRs~\cite{dsac,dsac*,ace2023,hscnet2020} training a separate model for each scene, in our experiments, we take each dataset as a whole and train a single model to recognize all landmarks in that dataset. The number of landmarks for different datasets are demonstrated in Table~\ref{tab:recognition}.

\textbf{Metric.} Following prior works~\cite{dsac,encn,sc-wls2022,lbr}, we report the median position and orientation errors ($cm, ^\circ$) and the success ratio of query images with pose errors within in ($5cm, 5^{\circ}$) for 7Scenes~\cite{sevenscenes2013} and 12Scenes~\cite{twelvescenes2016} datasets. For CambridgeLandmarks~\cite{posenet}, in addition to the median position and orientation errors, we also provide the success ratio at error threshold of ($0.25m, 2^\circ$). For Aachen dataset~\cite{aachen}, we use the official metric\footnote{https://www.visuallocalization.net/} by providing the success ratio of poses within error thresholds of ($0.25m, 2^\circ$), ($0.5m, 5^\circ$), and ($5m, 10^\circ$), respectively. Additionally, as efficiency is also important to real applications, we also analyze the time and memory efficiency of all methods at test time.

\textbf{Baseline.} We compare \method with previous state-of-the-art APRs~\cite{posenet,glnet,lsg,atloc2020,mapnet,mstrasformer2021,pae2022,lens2022}, SCRs~\cite{dsac,dsac*,ace2023,hscnet2020,kfnet2020,vsnet2021,nerfloc2023,camnet2019,sanet2019,sc-wls2022}, and HMs~\cite{as,csl,cpf,r2d2,d2-net,hfnet,sfd22023,ssm,vls, smc,backtofeature,nerfmatch,vrsnerf2024}. Results of SFD2+IMP and SP+SG are obtained from officially released source code with NetVLAD providing 20, 20, 20, and 50 candidate reference images on 7Scenes, 12Scenes, CambridgeLandmarks, and Aachen datasets, respectively. Results of other methods are from their papers or officially released source code.

\section{Experiments}
\label{sec:experiments}

In this section, we first present landmark generation and recognition results in Sec.~\ref{sec:exp:rec}. Then, we discuss the localization results, map size, and running time in Sec.~\ref{sec:exp:pose}, Sec.~\ref{sec:exp:map}, and Sec.~\ref{sec:exp:time}, respectively. Finally, we conduct an ablation study to verify the efficacy of different components in \method in Sec.~\ref{sec:exp:ablation} and visualize qualitative results in Sec.~\ref{sec:exp:qual_results}.

\setlength{\tabcolsep}{5.pt}

\begin{table}[t]
	\centering
	\caption{\textbf{Recognition results.} For 7Scenes~\cite{sevenscenes2013} (S) and 12Scenes~\cite{twelvescenes2016} (T) datasets, we assign each scene with 16 landmarks and obtain 112 = 7 × 16 and 192 = 12 × 16 landmarks in total, respectively. Each scene in CambridgeLandmarks~\cite{posenet} (C) dataset has 32 landmarks, so this
dataset has 160 = 5 × 32 landmarks in total. Aachen~\cite{aachen} (A) has only one scene with 512 landmarks. We train a model to recognize all landmarks in each dataset and report the recognition precision of top@1. The number of scenes (\#Scenes), landmarks (\#Landmarks), and the average number of keypoints (\#keypoints) are also included for reference. NA indicates no testing images.}
		\begin{tabular}{lcc|c|c}
			\toprule
			Dataset & \#Scenes & \#Landmarks & \#Keypoints & Precision@1 \\
			& &  & day, night & day, night\\
			\midrule
			S~\cite{sevenscenes2013} & 7 & 112 & 412.7, NA & 80.0\%, NA \\
			T~\cite{twelvescenes2016} & 12 & 192 & 435.3, NA & 95.0\%, NA \\
			C~\cite{posenet} & 5 & 160 & 4087.8, NA & 87.6\%, NA \\
			A~\cite{aachen} & 1 & 512 & 3192.2, 3262.7 & 69.8\%, 49.7\% \\
			\bottomrule
		\end{tabular}
\label{tab:recognition}
\end{table}

\subsection{Landmark generation and recognition}
\label{sec:exp:rec}

Fig.~\ref{fig:map_3d_seg} shows the map represented by 3D landmarks of 7Scenes~\cite{sevenscenes2013}, 12Scenes~\cite{twelvescenes2016}, CambridgeLandmarks~\cite{posenet}, and Aachen~\cite{aachen} datasets. The self-supervised landmark definition allows us to generate landmarks in any place in indoor (pumpkin in 7Scenes, apt1/kitchen in 12Scenes), outdoor (Kings College and Marys Church in CambridgeLandmarks), and even city-scale (Aachen) scenes. This technique breaks the limitation of using classic object classes such as buildings, trees, and so on as landmarks because different objects such as ovens and washing machines in indoor scenes can be easily merged as a single landmark based on their spatial connections (apt1/kitchen/12Scenes) and different parts of a building from outdoor scenes can be divided into different landmarks (Marys Church/CambdridgeLandmarks).

As shown in Table~\ref{tab:recognition}, for the evaluation of recognition performance, we report the average top@1 precision ($\frac{\#correct\ keypoints}{\#all\ keypoints}$) in each dataset for day and night images, separately. Since ground-truth poses of query images in Aachen dataset~\cite{aachen} are not publicly available, we use poses given by SFD2+IMP~\cite{sfd22023,imp2023} to generate pseudo ground-truth labels for sparse keypoints of query images. Note that although SFD2+IMP~\cite{sfd22023,imp2023} achieves state-of-the-art accuracy on the Aachen dataset~\cite{aachen}, estimated poses especially on night images are not perfect, resulting in errors in the ground-truth of recognition results.

\setlength{\tabcolsep}{3.pt}

\begin{table}[t]
	\centering
	\caption{\textbf{Localization accuracy on 7Scenes~\cite{sevenscenes2013} and 12Scenes~\cite{twelvescenes2016} datasets}. We report the median position ($cm$) and orientation ($\circ$) errors and localization precision with position and orientation errors within ($5cm,5^\circ$). NA indicates values unavailable.}
		\begin{tabular}{llc|c}
			\toprule
			Group & Method & 7Scenes & 12Scenes \\
			& & \multicolumn{2}{c}{$t (cm)\downarrow/R (^\circ)\downarrow/Percent (5cm, 5^\circ$)$\uparrow$} \\
			\toprule
			\multirow{8}{*}{APRs}
 		
			&LsG~\cite{lsg} & 19 / 7.5 / NA & NA / NA / NA  \\
			&AtLoc~\cite{atloc2020} & 20 / 7.6 / NA & NA / NA / NA  \\
			&PAEs~\cite{pae2022} & 19 / 7.5 / NA & NA / NA / NA  \\
			&LENS~\cite{lens2022} & 8 / 3.0 / NA & NA / NA / NA  \\
            &DFNet~\cite{dfnet2022} & 12 / 3.7 / NA & NA / NA/ NA\\\
			&Posenet~\cite{posenet} & 24 / 7.9 / 1.9 & NA / NA / NA \\
			&MapNet~\cite{mapnet} & 21 / 7.8 / 4.9 & NA / NA / NA  \\
			&MS-Transformer~\cite{mstrasformer2021} & 18 / 7.3 / 4.1 & NA / NA / NA  \\
			&GLNet~\cite{glnet} & 19 / 6.3 / 7.8 & NA / NA / NA \\
            & NeFeS~\cite{nefes2024} & 2 / 0.8 / 78.3 & NA / NA / NA \\

			\midrule
			\multirow{7}{*}{SCRs}
			& VSNet~\cite{vsnet2021} & 2.4 / 0.8 / NA & NA / NA / NA  \\
			& PixLoc~\cite{backtofeature} & 2.9 / 1.0 / NA & NA / NA / NA \\
		& CAMNet~\cite{camnet2019} & 4 / 1.7 / NA & NA / NA / NA \\
		& SANet~\cite{sanet2019} & 5.1 / 1.7 / NA & NA / NA / NA \\
        &SRC~\cite{src2022} & 4.7 / 1.6 / NA & NA / NA / NA \\
        &DSM~\cite{dsm2021} & 2.7 / 0.9 / 81.5 & NA / NA / NA\\
        &SC-wLS~\cite{sc-wls2022} & 7 / 1.5 / 43.2 & NA / NA / NA  \\
        &HSCNet~\cite{hscnet2020} & 3 / 0.9 / 84.8 & 0.1 / 0.5 / 99.3 \\
		& KFNet~\cite{kfnet2020} & 2.9 / 1.0 / NA & NA / NA / 98.9 \\
		&NeRF-loc~\cite{nerfloc2023} & 2.3 / 1.3 / 89.5 & NA / NA / NA  \\
		
		&DSAC*~\cite{dsac*} & 2.7 / 1.4 / 96.0 & NA / NA / 99.6 \\
		&ACE~\cite{ace2023} & 2.5 / 1.3 / 97.1 & 1.0 / 0.4 / 99.9 \\

			\midrule
			\multirow{3}{*}{HMs}
			&AS~\cite{as} & NA / NA / 98.5 & NA / NA / 99.8 \\ 
            & NeRFMatch~\cite{nerfmatch} & 2.8 /0.7 / 78.4 & NA / NA / NA \\
            & VRS-NeRF~\cite{vrsnerf2024} & 1.0 / 0.3 / 93.1 & NA / NA / NA \\
			& SP+SG~\cite{superpoint,superglue} & 1.0 / 0.2 / 95.7 & 1.0 / 0.1 / 100 \\
			& SFD2+IMP~\cite{sfd22023,imp2023} & 1.0 / 0.1 / 95.7 & 1.0 / 0.1 / 99.7\\
			
			\midrule
			&\textbf{Ours} & 1.0 / 0.3 / 97.3 & 1.0 / 0.1 / 97.8 \\
			\bottomrule
		\end{tabular}
\label{tab:7scenes}
\end{table}

\setlength{\tabcolsep}{6pt}

\begin{table*}[t]
	\centering
	\caption{\textbf{Localization accuracy on the CambridgeLandmarks dataset~\cite{posenet}.} We report the median position (cm), orientation ($^\circ$) errors and localization success ratio of pose errors within ($25cm, 2^\circ$). As many methods are not evaluated on Great Court, the results on Great Court are only used as reference. NA indicates values unavailable.}
		\begin{tabular}{llccccc|c}
			\toprule
			Group & Method & Kings College  & Old Hospital & Shop Facade & StMarys Church & Average &  Great Court\\
			& & \multicolumn{6}{c}{$t (cm)\downarrow/R (^\circ)\downarrow/Percent (25cm, 2^\circ$)$\uparrow$} \\
			\midrule
			
			\multirow{5}{*}{APRs} & MapNet~\cite{mapnet} & 107 / 1.9 / NA  & 149 / 4.2 / NA & 200 / 4.5 / NA & 194 / 3.9 / NA & 163 / 3.6 / NA & 785 / 3.8 / NA\\
			& PAEs~\cite{pae2022} & 90 / 1.5 / NA  &  207 / 2.6 / NA & 99 / 3.9 / NA & 164 / 4.2 / NA &  140 / 3.1 / NA & NA / NA / NA\\
			& LENS~\cite{lens2022} & 33 / 0.5 / NA & 44 / 0.9 / NA & 27 / 1.6 / NA & 53 / 1.6 / NA & 39 / 1.2 / NA & NA / NA / NA \\	
            &DFNet~\cite{dfnet2022} & 43 / 0.9 / NA & 46 / 0.9 / NA & 16 / 0.6 / NA & 50 / 1.5 / NA & 39 / 1.0 / NA & NA / NA / NA \\
			& Posenet~\cite{posenet} & 88 / 1.0 / 0 & 88 / 3.8 / 0 & 157 / 3.3 / 0 & 320 / 3.3 / 0 & 163 / 2.9 / 0 & 683 / 3.5 / 0 \\
			&MS-Transformer~\cite{mstrasformer2021} & 83 / 1.5 / 3.5   &  181 / 2.4 / 2.2  & 86 / 3.1 / 4.9  & 162 / 4.0 / 0.4 & 128 / 2.8 / 2.8 & NA / NA / NA\\
			&GLNet~\cite{glnet} & 59 / 0.7 / 3.4  & 50 / 2.9 / 3.6 & 190 / 3.3 / 7.9 & 188 / 2.8 / 0.6 & 122 / 2.4 / 3.9 & NA / NA / NA\\
            & NeFeS~\cite{nefes2024} & 37 / 0.5 / 30.6  &  52 / 0.9 / 18.1 & 15 / 0.5 / 70.9 & 37 / 1.1 / 33.6 & 35 / 0.8 / 38.3 & NA / NA / NA\\
						
			\midrule
			\multirow{3}{*}{SCRs}
			& NeRF-loc~\cite{nerfloc2023} & 7 / 0.2 / NA  &  18 / 0.4 / NA & 11 / 0.2 / NA &  4 / 0.2 / NA & 10 / 0.3 / NA & 25 / 0.1 / NA\\
			& HSCNet~\cite{hscnet2020} & 18 / 0.3 / NA  & 19 / 0.3 / NA & 6 / 0.3 / NA & 9 / 0.3 / NA & 13 / 0.3 / NA & 28 / 0.2 / NA\\
            & SRC~\cite{src2022} & 39 / 0.7 / NA  & 38 / 0.5 / NA & 19 / 1.0 / NA & 31 / 1.0 / NA & 32 / 0.8 / NA & 81 / 0.5 / NA\\
            & DSM~\cite{dsm2021} & 19 / 0.4 / NA  & 23 / 0.4 / NA & 6 / 0.3 / NA & 11 / 0.3 / NA & 15 / 0.4 / NA & 43 / 0.2 / NA\\
             & NeuMap~\cite{neumap2023} &  14 / 0.2 / NA & 19 / 0.4 / NA & 6 / 0.3 / NA & 17 / 0.5 / NA & 14 / 0.4 / NA  & 6 / 0.1 / NA\\
			& SC-WLS~\cite{sc-wls2022} &  14 / 0.6 / 68.2  & 42 / 1.7 / 23.1 &  11 / 0.7 / 76.7 & 39 / 1.3 / 34.0 & 27 / 1.1 / 50.5 & 164 / 0.9 / 7.3\\
            
			& DSAC*~\cite{dsac*} & 13 / 0.4 / 72.3  & 20 / 0.3 / 57.1 & 6 / 0.3 / 91.3 & 13 / 0.4 / 81.5 & 13 / 0.4 / 75.6 & 40 / 0.2 / 22.2\\	
			
			& ACE~\cite{ace2023} & 28 / 0.4 / 44.6  & 31 / 0.6 / 40.7 & 5 / 0.3 / 98.1 & 19 / 0.6 / 61.3 & 21 / 0.5 / 61.2  &  42 / 0.2 / 28.7 \\
			
			\midrule
			\multirow{3}{*}{HMs} & AS~\cite{as} & 24 / 0.1 / NA & 20 / 0.4 / NA & 4 / 0.2 / NA & 8 / 0.3 / NA & 14 / 0.3 / NA & 13 / 0.2 / NA \\	
			& PixLoc~\cite{backtofeature} & 14 / 0.2 / NA  &  16 / 0.3 / NA &  5 / 0.2 / NA & 10 / 0.3 / NA & 11 / 0.3 / NA &  30 / 0.1 / NA\\	
			& SANet~\cite{sanet2019} &  32 / 0.5 / NA  &  32 / 0.5 / NA &  10 / 0.5 / NA & 16 / 0.6 / NA & 23 / 0.5 / NA &  328 / 2.0 / NA\\
             & CrossFire~\cite{crossfire2023} & 47 / 0.7 / NA & 43 / 0.7 / NA & 20 / 1.2 / NA & 39 / 1.4 / NA & 37 / 1.0 / NA & NA / NA / NA \\ 
           
            & InLoc~\cite{inloc2018} & 46 / 0.8 / NA & 48 / 1.0 / NA & 11 / 0.5 / NA & 18 / 0.6 / NA & 31 / 0.7 / NA  & 120 / 0.6 / NA \\
           
            & NeRFMatch~\cite{nerfmatch} & 13 / 0.2 / NA  & 21 / 0.4 / NA & 9 / 0.4 / NA & 11 / 0.4 / NA & 14 / 0.4 / NA & 20 / 0.1 / NA\\
            
            & VRS-NeRF~\cite{vrsnerf2024} & 9 / 0.1 / 90.7   & 11 / 0.2 / 79.7 &  2 / 0.1 / 96.1 & 5 / 0.2 / 90.8 & 7 / 0.2 / 89.3 & NA / NA / NA\\
			& SFD2+IMP~\cite{sfd22023,imp2023} & 7 / 0.1 / 94.8  & 10 / 0.2 / 80.2 & 2 / 0.1 / 97.1 & 4 / 0.1 / 98.7 & 6 / 0.1 / 92.7 & 11 / 0.1 / 74.6\\
			& SP+SG~\cite{superpoint,superglue} & 7 / 0.1 / 94.5 & 9 / 0.2 / 81.9 & 2 / 0.1 / 97.1 & 4 / 0.1 / 98.7 & 6 / 0.1 / 93.0 & 12 / 0.1 / 74.7 \\  
			
			\midrule
			& \textbf{Ours} & 15 / 0.1 / 71.1  & 9 / 0.2 / 78.0 & 2 / 0.1 / 98.1 & 5 / 0.2 / 96.8 & 8 / 0.2 / 86.0 & 16 / 0.1 / 62.5\\
			\bottomrule
		\end{tabular}

\label{tab:cambridgelandmarks}
\end{table*}

Table~\ref{tab:recognition} shows that in indoor scenes such as 7Scenes and 12Scenes datasets~\cite{sevenscenes2013, twelvescenes2016} consisting of 112 and 192 landmarks respectively, \method gives precision over 80\% by using about 400 keypoints for each query image. \method yields around 87.9\% precision on the CambridgeLandmarks~\cite{posenet} with approximate 4,000 keypoints as input. For the Aachen dataset~\cite{aachen}, the precision degrades to 69.8\% due to large viewpoint changes of day images. Illumination changes and the missing night images for training further impair the performance, leading to the precision of 49.7\% on night images. Although the top@1 precision of night images is not as high as that of day images, our strategy of retaining several candidate landmarks for progressive verification as introduced in Sec.~\ref{sec:method:sparse_rec} increases the localization success ratio. In the future, more data augmentation could be adopted to improve the recognition performance on images with large viewpoint and illumination variations.

\subsection{Localization results}
\label{sec:exp:pose}

\textbf{7Scenes (S) and 12Scenes (T).} Table~\ref{tab:7scenes} shows the average median position and orientation errors and the success ratio at the error threshold of ($5cm,5^\circ$) on the 7Scenes~\cite{sevenscenes2013} and 12Scenes~\cite{twelvescenes2016} datasets. From Table~\ref{tab:7scenes}, we can see that APRs have the largest position and orientation errors due to implicit 3D information embedding and their similar behavior to image retrieval~\cite{sattler2019understanding}. Among APRs, by performing pose refinement, NeFeS~\cite{nefes2024} gives the best performance with close median errors to SCRs. However, its accuracy at the error threshold of ($5cm,5^\circ$) is still not comparable to the best methods of SCRs (78.3\% vs. 97.1\%) and HMs (78.3\% vs. 98.5\%).

With explicit 2D-3D coorespondences, SCRs achieve excellent median position ($\approx3cm$) and orientation ($\approx1^\circ$) errors around 7$\times$ smaller than APRs ($\approx20cm, 7^\circ$). Note that each scene in 7Scenes~\cite{sevenscenes2013} and 12Scenes~\cite{twelvescenes2016} datasets has a size of only $2m\times2m\times2m$, so SCRs are able to regress 3D coordinates accurately for precise pose estimation. Table~\ref{tab:cambridgelandmarks} demonstrates when the scales of scenes increase, their accuracy drops significantly.

HMs obtain close accuracy to SCRs. Among HMs, AS~\cite{as} (98.5\% and 99.8\%) works slightly better than SP+SG~\cite{superpoint,superglue} (95.7\% and 100\%) and SFD2+IMP~~\cite{sfd22023,imp2023} (95.7\% and 99.7\%), because AS extracts multi-scale SIFT~\cite{sift} as local features which are slightly robust to areas with similar structures (e.g., stairs in 7Scenes) than single-scale corners used by SP~\cite{superpoint} and SFD2~\cite{sfd22023}. NeRFMatch~\cite{nerfmatch} uses NeRFs for implicit map representation and produces relatively worse accuracy. VRS-NeRF~\cite{vrsnerf2024} trains a NeRF model for each scene and performs patch-wise rendering, so it gives better performance than NeRFMatch. 

\method ($1cm,0.3^\circ$) gives close numbers to SP+SG~\cite{superpoint,superglue} ($1cm,0.1^\circ$) and SFD2+IMP\cite{sfd22023,imp2023} ($1cm,0.1^\circ$) in terms of median errors. Benefitting from the semantic-wise matching which is more robust to similar structures (e.g., stairs in 7Scenes), \method (97.3\%) produces slightly higher accuracy on 7Scenes dataset. Because of insufficient SFD2 features for recognition and 2D-3D matching in textureless regions (e.g., office2/5a in 12Scenes), \method reports slightly worse accuracy on 12Scenes~\cite{twelvescenes2016} (97.8\%). By searching for 20 candidate images to build correspondences, SP+SG and SFD2+IMP partially solve this problem.

\setlength{\tabcolsep}{3.5pt}

\begin{table}[t]
	\centering
	\caption{\textbf{Results on Aachen dataset~\cite{aachen,visbenchmark}.} We report the success ratio at error thresholds of $(0.25m, 2^\circ)$, $(0.5m, 5^\circ)$, and $(5m, 10^\circ)$. * indicates methods using semantics.}
	\begin{tabular}{llcc}
		\toprule
		Group & Method & Day & Night\\
		& & \multicolumn{2}{c}{$(0.25m, 2^\circ)/(0.5m, 5^\circ)/(5m, 10^\circ)\uparrow$} \\	
		\midrule
		
		\multirow{3}{*}{SCRs}
		& ESAC~\cite{esac2019} & 42.6 / 59.6 / 75.5 & 3.1 / 9.2 / 11.2 \\
		& HSCNet~\cite{hscnet2020} & 71.1 / 81.9 / 91.7 & 32.7 / 43.9 / 65.3 \\
		& NeuMap~\cite{neumap2023} & 80.8 / 90.9 / 95.6 & 48.0 / 67.3 / 87.8 \\
		\midrule
		
		\multirow{15}{*}{HMs} & SSM*~\cite{ssm} & 71.8 / 91.5 / 96.8 & 58.2 / 76.5 / 90.8 \\
		& VLM*~\cite{lln} & 62.4 / 71.8 / 79.9 & 35.7 / 44.9 / 54.1 \\
		& SMC*~\cite{smc} & 	52.3 / 80.0 / 94.3 & 29.6 / 40.8 / 56.1\\
		& LBR*~\cite{lbr} & 88.3 / 95.6 / 98.8 & 84.7 / 93.9 / 100.0  \\
		& SFD2*~\cite{sfd22023} & 88.2 / 96.0 / 98.7 & 87.8 / 94.9 / 100.0 \\
		
		& AS ~\cite{as} & 85.3 / 92.2 / 97.9 & 39.8 / 49.0 / 64.3 \\
		& CSL~\cite{csl} & 52.3 / 80.0 / 94.3 & 29.6 / 40.8 / 56.1 \\
		& CPF~\cite{cascadedfiltering} &  76.7 / 88.6 / 95.8 & 33.7 / 48.0 / 62.2  \\
		& SceneSqqueezer~\cite{scenesqueezer2022} & 75.5 / 89.7 / 96.2 & 50.0 / 67.3 / 78.6 \\
		& SP~\cite{superpoint} & 80.5 / 87.4 / 94.2 & 42.9 / 62.2 / 76.5 \\
		& R2D2~\cite{r2d2} & NA / NA / NA & 76.5 / 90.8 / 100.0 \\
		& ASLFeat~\cite{aslfeat} & NA / NA / NA & 81.6 / 87.8 / 100.0 \\
		& D2Net~\cite{d2-net} & 84.8 / 92.6 / 97.5 &  84.7 / 90.8 / 96.9 \\
        & VRS-NeRF~\cite{vrsnerf2024} & 70.1 / 76.9 / 80.9 & 44.9 / 51.0 / 62.2 \\
		& SP+SG~\cite{superpoint, superglue} & 89.6 / 95.4 / 98.8 & 86.7 / 93.9 / 100.0 \\
		& SFD2+IMP~\cite{sfd22023,imp2023} & 89.7 / 96.5 / 98.9	& 84.7 / 94.9 / 100.0 \\
		\midrule
		
		& \textbf{Ours} & 	82.5 / 90.9 / 96.7	& 77.6 / 87.8 / 94.9 \\
		\bottomrule			   
	\end{tabular}
	\label{tab:aachen}
\end{table}


\textbf{CambridgeLandmarks (C).} Table~\ref{tab:cambridgelandmarks} shows the results on CambridgeLandmarks~\cite{posenet} in terms of median position (cm) and orientation errors ($^\circ$) and the pose success ratio at the error threshold of ($25cm, 2^\circ$). Table~\ref{tab:cambridgelandmarks} demonstrates that APRs~\cite{posenet,lsg,glnet} give very large errors on average, especially on position estimation ($\approx35cm$). NeFeS~\cite{nefes2024} also gives the best performance among APRs by adopting pose refinement. However, its success ratio is about 37\% and 55\% lower than the best method of APRs (38.3\% vs. 75.5) and HMs (38.3\% vs. 93.0\%). Although SCRs~\cite{dsac,dsac*,ace2023,vsnet2021,hscnet2020} achieve outstanding median position ($\approx20cm$) and orientation ($\approx0.3^\circ$) errors, their success ratios at the error threshold of $(25cm, 2^\circ)$ are not satisfying. Even the state-of-the-art DASC*~\cite{ace2023} has only about 75.6\% on average, which is about 17\% lower than the best of HMs~\cite{sfd22023,superpoint} (93.0\%). This observation reveals the limitations of SCRs in large-scale scenes.

HMs give the best accuracy. With more robust local features and advanced matchers, SFD2+IMP ($6cm,0.1^\circ$) and SP+SG ($6cm,0.1^\circ$) work better than others including PixLoc ($11cm,0.3^\circ$) and AS ($14cm,0.3^\circ$) on median position and rotation errors. By training a model for each scene and adopting IMP~\cite{imp2023} for matching, VRS-NeRF~\cite{vrsnerf2024} reports close accuracy to SP+SG~\cite{superpoint,superglue} and SFD2+IMP~\cite{sfd22023,imp2023}. On average, \method (86.0\%) yields about 48\% and 10\% higher accuracy than the best of ARPs (38.3\%) and SCRs (75.6\%) and around 34\% on the Great Court (62.5\% vs. 28.7\%). Besides, \method also gives competitive accuracy to SFD2+IMP (92.7\%) and SP+SG (93.0\%). The around 7\% lower accuracy (86.0\% vs. 93.0\%) mainly comes from the scene of Kings College due to high uncertainties of 3D points with large depth values. Moreover, \method takes all 5 scenes as a whole and performs recognition and registration in the whole dataset as shown in Table~\ref{tab:recognition}, making the localization more challenging. Conversely, APRs, SCRs, and HMs do localization in each scene of the CambridgeLandmarks, separately.


\setlength{\tabcolsep}{6pt}

\begin{table*}[t]
	\centering
	\caption{\textbf{Map size.} We show the number of reference image (\#Ref. image), 3D points (\#3D point) in million, local descriptor size (local desc.) in gigabyte, global descriptor size (global desc.) in gigabyte, 3D map size in gigabyte, and the total size at test time. We also provide the ratio of total size to that of SFD2+IMP on 7Scenes (S)~\cite{sevenscenes2013}, 12Scenes (T)~\cite{twelvescenes2016}, CambridgeLandmarks (C)~\cite{posenet} and Aachen (A)~\cite{aachen}. \textcolor{red}{Red} and \textcolor{green}{green} indicate the total size larger or not than that of SFD2+IMP.}
		\begin{tabular}{llccccc|c}
			\toprule
			 Dataset & Method & \#Ref. image & \#3D point (M) & Local desc. (G) & Global desc. (G) & Map size (G) & Total size (Ratio) \\
			\midrule
			\multirow{4}{*}{S} 
			&Posenet~\cite{posenet} & 0 & 0 & 0 & 0 & 0.33 & 0.33 \textcolor{green}{(4.7\%)} \\
			& DSAC*~\cite{dsac*} & 0 & 0 & 0 & 0 & 0.03 & 0.03 \textcolor{green}{(0.4\%)}\\
			& ACE~\cite{ace2023} & 0 & 0 & 0 & 0 & 0.004 & 0.004 \textcolor{green}{(0.04\%)}\\
			
			& SP+SG~\cite{superpoint,superglue,netvlad} & 26,000 & 0.66 & 14.6 & 0.79 & 0.50 & 15.90 \textcolor{red}{(229.3\%)}\\
			& SFD2+IMP~\cite{sfd22023,imp2023} & 26,000 & 0.44  & 5.78 & 0.79 & 0.39 & 6.96 \textcolor{green}{(1.0\%)}\\ 
			
			& \textbf{Ours} & 0 & 0.05 & 0 & 0 & 0.03 & 0.03 \textcolor{green}{(0.4\%)} \\ 
			\midrule
			
			\multirow{4}{*}{T} 
			&Posenet~\cite{posenet} & 0 & 0 & 0 & 0 & 0.56 & 0.56 \textcolor{green}{(10.6\%)}\\
			& DSAC*~\cite{dsac*} & 0 & 0 & 0 & 0 & 0.03 & 0.03 \textcolor{green}{(0.3\%)} \\
			& ACE~\cite{ace2023} & 0 & 0 & 0 & 0 & 0.004 & 0.004 \textcolor{green}{(0.08\%)}\\
			& SP+SG~\cite{superpoint,superglue,netvlad} & 16,989 & 0.84 & 11.62 & 0.52 &  0.43 & 12.57 \textcolor{red}{(237.2\%)}\\
			& SFD2+IMP~\cite{sfd22023,imp2023} & 16,989 & 0.59 & 4.45 & 0.52 & 0.33 & 5.30 \textcolor{green}{(1.0\%)}\\ 
		
			& \textbf{Ours} & 0 & 0.15 & 0 & 0 & 0.08 & 0.08 \textcolor{green}{(1.5\%)}\\ 
			\midrule
			
			\multirow{4}{*}{C} 
			&Posenet~\cite{posenet} & 0 & 0 & 0 & 0 & 0.23 & 0.23 \textcolor{green}{(3.8\%)} \\
			& DSAC*~\cite{dsac*} & 0 & 0 & 0 & 0 & 0.14 & 0.14 \textcolor{green}{(2.3\%)}\\
			& ACE~\cite{ace2023} & 0 & 0 & 0 & 0 & 0.02 & 0.02 \textcolor{green}{(0.3\%)}\\
			& SP+SG~\cite{superpoint,superglue,netvlad} & 5,365 & 1.65 & 10.81 & 0.16 & 0.63 & 11.60 \textcolor{red}{(190.2\%)}\\
			& SFD2+IMP~\cite{sfd22023,imp2023} & 5,365 & 1.55 & 5.31 & 0.16 & 0.63 & 6.10 \textcolor{green}{(1.0\%)}\\

			& \textbf{Ours}& 0 & 0.4 & 0 & 0 & 0.22 & 0.22 \textcolor{green}{(3.6\%)}\\ 
			\midrule
			
			\multirow{4}{*}{A} & SP+SG~\cite{superpoint,superglue,netvlad} & 6,697 & 2.15 & 13.83 & 0.20 & 0.72 & 14.76 \textcolor{red}{(180.7\%)}\\
			& SFD2+IMP~\cite{sfd22023,imp2023} & 6,697 & 2.21 & 7.21 & 0.20 & 0.80 & 8.17 \textcolor{green}{(1.0\%)}\\ 
			& \textbf{Ours}& 0 & 1.33 &  0 & 0 & 0.72 & 0.72 \textcolor{green}{(8.8\%)} \\ 
			
			
			\bottomrule
		\end{tabular}
\label{tab:map_size}
\end{table*}

\textbf{Aachen (A).} Table~\ref{tab:aachen} shows results on the Aachen dataset~\cite{aachen}. It is not surprising that SCRs including ESAC~\cite{esac2019} (3.1\%), HSCNet~\cite{hscnet2020}  (23.7\%), and NeuMap~\cite{neumap2023} (48.0\%) report relatively poor accuracy especially on night images because Aachen dataset has much larger scale than CambridgeLandmarks~\cite{posenet}, 7Scenes~\cite{sevenscenes2013}, and 12Scenes~\cite{twelvescenes2016} datasets and is more challenging due to appearance changes. HMs achieve state-of-the-art accuracy. Among HMs, AS~\cite{as} gives relatively worse performance partially because of the less robustness of SIFT~\cite{sift} features to appearance changes. Because of the limited representation ability of NeRFs in city-scale environments and insufficient training data, VRS-NeRF~\cite{vrsnerf2024} gives very poor accuracy even if the advanced IMP matcher~\cite{imp2023} is incorporated. SFD2+IMP~\cite{sfd22023,imp2023} obtains the best performance despite having a smaller descriptor size than SP~\cite{superpoint}. With the assistance of semantics, semantic-aware methods including LBR~\cite{lbr} also report promising performance. However, they utilize the standard framework of HMs to perform retrieval and 2D-2D matching, so they have low memory and time efficiency.

\method outperforms SCRs especially on night images (77.6\% vs. 48.0\%) as our landmark recognition is more robust than pixel-wise regression in large-scale scenes. Compared with state-of-the-art HMs such as SFD2+IMP and SP+SG, \method also gives competitive accuracy that is about 7\% lower for both day (82.5\% vs. 89.7\%) and night (77.6\% vs. 84.7\%) images. The lower accuracy partially comes from the limited training data of sparse reference images for recognition.  As \method is the first method performing localization through 3D landmark recognition, we believe that it can be further improved by potential strategies discussed in Sec.~\ref{sec:conclusion_futurework}.


\subsection{Map size}
\label{sec:exp:map}

Table~\ref{tab:map_size} shows the map size of APRs~\cite{posenet,lsg}, SCRs~\cite{dsac*,ace2023}, HMs~\cite{superpoint,superglue,sfd22023,imp2023}, and \method. APRs and SCRs are end-to-end frameworks, so we use their model size as the map size for reference only and mainly focus on the comparison between HMs and \method. We report the number of reference images (\#Ref. image), 3D points in million (\#3D point), local descriptor size (Local desc.), global descriptor size (Global desc.), 3D map size (Map size), and the total size. Note that for HMs, invalid local keypoints are removed and only local keypoints with 3D correspondences are considered. Obviously, SP+SG~\cite{superpoint,superglue} has the largest size because of the local and global descriptors. By using a smaller dimension (128) of local descriptors than SP~\cite{superpoint} (256), SFD2+IMP~\cite{sfd22023,imp2023} reduces about 50\% map size while preserving localization accuracy as shown in Table~\ref{tab:7scenes},~\ref{tab:cambridgelandmarks}, and~\ref{tab:aachen}. 

Compared with SFD2+IMP on the Aachen dataset, PRAM saves about 2.5\% (0.20G) and 88\% (7.21G) of the total size (8.17G) by discarding global and local descriptors, respectively. With the introduced virtual reference frames and the adaptive landmark-wise 3D point pruning strategy, PRAM reduces about 40\% redundant 3D points (1.33 million vs. 2.21 million) and additionally saves about 0.2\% (0.72G vs. 0.80G) of map size. As a result, PRAM reduces over 90\% map size of SFD2+IMP on the Aachen dataset and even more on other datasets. This is very important to applications on devices with limited computing resources.

\subsection{Running time}
\label{sec:exp:time}

We test the average time of APRs, SCRs, HMs, and \method for processing one image in Marys Church of CambridgeLandmarks~\cite{posenet} on a machine with RTX3090 GPU and Xeon(R) Silver 4216 CPU@2.10GHz. For a fair comparison, we use the same image resolution, number of keypoints, and reconstruction framework~\cite{hfnet} for all methods. ResNet34~\cite{resnet} is adopted as the backbone of APRs and fastest ACE~\cite{ace2023} is adopted as the representative of SCRs.

\setlength{\tabcolsep}{2.0pt}

\begin{table}[t]
	\centering
	\caption{\textbf{Running time (ms).} We show the average time of processing one frame in CambridgeLandmarks~\cite{posenet}. The time (ms) of local feature extraction (Local), global feature extraction (Global), recognition (Rec.), tracking (Track.), and pose estimation (Pose) are included. The total processing time is also provided.}
		\begin{tabular}{lccccc|c}
			\toprule
			Method & Local & Global & Rec. & Track. & Pose & Total \\
			\midrule
			APRs~\cite{posenet,atloc2020,nefes2024} & 40 & 0 & 0 & 0 & 0 & 40\\
			SCRs~\cite{ace2023} & 10 & 0 & 0 & 0 & 60 & 70 \\
			SP+SG+NV~\cite{superpoint,superglue,netvlad} & 50 & 80 & 0 & 0 & 3690 & 3820\\
			SFD2+IMP+NV~\cite{sfd22023,imp2023,netvlad} & 60 & 80 & 0 & 0 & 2520 & 2660\\
			\midrule
			\textbf{Ours (tracking)} & 60 & 0 & 20 & 310 & 0 & 390 \\
			\textbf{Ours} & 60 & 0 & 20 & 310 & 730 & 1120 \\
			\bottomrule
		\end{tabular}
\label{tab:time}
\end{table}

Table~\ref{tab:time} shows the time of local feature~\cite{superpoint,sfd22023} extraction (Local), global feature~\cite{netvlad} extraction (Global), recognition (Rec.), tracking (Track.), and pose estimation (Pose). APRs~\cite{posenet,lsg,atloc2020} are the fastest methods with local feature extraction as the only source of time cost. SCRs adopt lighter backbones, so they are faster on local feature extraction than APRs. However, SCRs~\cite{ace2023} work slower overall due to the pose estimation with RANSAC~\cite{ransac1981} on dense 2D-3D correspondences. For HMs, the time cost comes from local and global feature extraction and the pose estimation particularly. Therefore, HMs are much slower than APRs and SCRs. 

As with HMs, \method also needs time to extract local features. Fortunately, the time of sparse recognition is 20ms, 4$\times$ faster than NV~\cite{netvlad} (80ms). The tracking component used for progressive verification takes about 310ms due to the graph-based matching. With the initial pose given by tracking, our pose refinement uses only 730ms to recover a more accurate pose, which is about 3.5$\times$ and 5.1$\times$ faster than SFD2+IMP+NV and SP+SG+NV, respectively. As a result, the total time of \method is 2.4$\times$ and 3.4$\times$ less than SFD2+IMP+NV and SP+SG+NV. The speed of \method can be further improved by removing the heavy pose refinement component and utilizing the pose provided by tracking as results. By doing that, \method uses 390ms to process each frame at the expense of only 2\% accuracy loss (as shown in Table~\ref{tab:ablation}).  Although \method is faster than HMs, \method still has a large space for improvements, compared with APRs and SCRs.

\subsection{Ablation study}
\label{sec:exp:ablation}

We test the influence of different components in \method including the combination of local keypoints and matchers (F+M), clustering method (C.M), sparsification (Spar.), tracking (Track.), pose refinement (Ref.), dimension of 3D descriptors (\#D), and the number of candidate landmarks (\#C.L.). Results on Marys Church in CambridgeLandmarks at the error threshold of $(25cm, 2^\circ)$ are shown in Table~\ref{tab:ablation}.

\setlength{\tabcolsep}{4.5pt}

\begin{table}[t]
	\centering
	\caption{\textbf{Ablation study.} We test the influence of different feature+matcher combinations (F+M), clustering method (C.M), sparsification (Spar.), tracking (Track.), refinement (Ref.), descriptor dimension (\#D), and the number of candidate landmarks (\#C.L.) to pose success ratio within the error threshold of $(25cm,2^\circ)$.}
		\begin{tabular}{lccccccc}
			\toprule
			F+M & C.M. & Spar. & Track. & Ref. & \#D & \#C.L. & Acc$\uparrow$  \\
			\midrule
			SFD2+IMP & HCluster & \xmark & \cmark & \xmark & 128 & 20 & 94.5\% \\
			SFD2+IMP & HCluster & \cmark & \cmark & \xmark & 128 & 20 &  94.3\%\\
			SFD2+IMP & HCluster & \cmark & \cmark & \cmark & 128 & 20 & 96.8\% \\
            SFD2+IMP & k-means++ & \cmark & \cmark & \cmark & 128 & 20 & 92.5\% \\
            SP+SG & HCluster & \cmark & \cmark & \cmark & 256 & 20 & 75.5\% \\
            
			\midrule
			SFD2+IMP & HCluster & \cmark & \cmark & \cmark & 64 & 20 & 94.9\% \\
			SFD2+IMP & HCluster & \cmark & \cmark & \cmark & 32 & 20 & 94.3\% \\
			SFD2+IMP & HCluster & \cmark & \cmark & \cmark & 0 & 20 & 15.8\% \\
			\midrule
			SFD2+IMP & HCluster & \cmark & \cmark & \cmark & 128 & 10 & 95.3\% \\
			SFD2+IMP & HCluster & \cmark & \cmark & \cmark & 128 & 5 & 95.1\%  \\
			SFD2+IMP & HCluster & \cmark & \cmark & \cmark & 128 & 1 & 91.7\% \\
			\bottomrule
			
		\end{tabular}
\label{tab:ablation}
\end{table}

\begin{figure*}[t]
	\centering
	\includegraphics[width=1.\linewidth]{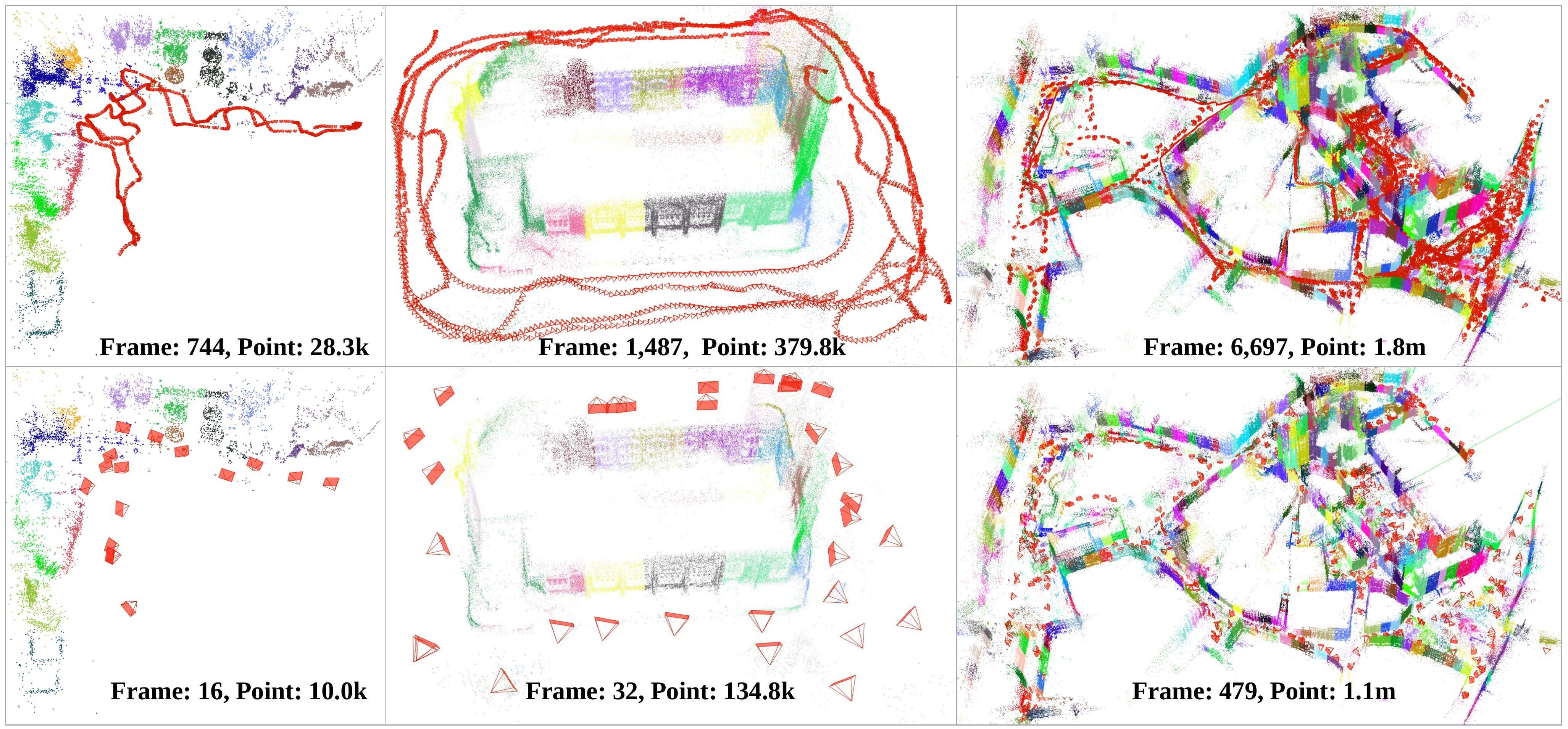}
	\caption[Sparse map of PRAM]{\textbf{Map sparsification.} Original (top) and sparsified maps (bottom) with reference frames and 3D points of apt1/kitchen in 12Scenes~\cite{twelvescenes2016} (left), StMarys Church in CambridgeLandmarks~\cite{posenet} (middle), and Aachen datasets~\cite{aachen} (right). The number of frames and 3D points from the original and the sparsified maps are included.
	}
	\label{fig:vis_map_sparsification}
\end{figure*}

According to Table~\ref{tab:ablation}, we have the following observations. (i) Sparsification has little influence on the accuracy (94.3\% vs. 94.5\%) as most discarded 3D points have high overlap with preserved ones. (ii) Refinement increases the accuracy by about 2.5\% (96.8\% vs. 94.3\%) because more 3D points are used to refine the pose in the refinement process. (iii) The hierarchical clustering works better than k-means++ because hierarchical clustering preserves the completeness of objects in the 3D space better than k-means++, resulting in higher recognition and localization performance; (iv) SFD2+IMP works better than SP+SG~\cite{superpoint,superglue} because SFD2 keypoints contain richer semantic information than SP keypoints and IMP also works better than SG. (v) When the dimension of 3D descriptors is reduced from 128 to 64 or 32, \method undergoes an acceptable loss of accuracy from 96.8\% to 94.9\% or 94.3\%, which means the map size can be further reduced by decreasing the descriptor size. However, when 3D descriptors are not involved in localization (only 2D/3D coordinates are used), the performance drops significantly from 94.3\% to 15.8\% compared with using 32-float descriptors because pure 2D/3D coordinates are not discriminative enough, especially for 2D-3D matching. (vi) When the number of candidate landmarks (\#C.L) used for tracking is reduced from 20 to 10, 5, and even 1, \method still gives a very robust performance with an approximate 5\% loss of accuracy (96.8\% vs. 91.7\%). Note that \method recognizes up to 160 landmarks in the whole CambridgeLandmarks~\cite{posenet}. This observation further proves the robustness of our sparse recognition strategy.

\subsection{Qualitative results}
\label{sec:exp:qual_results}  

\textbf{Map sparsification.} Fig.~\ref{fig:vis_map_sparsification} illustrates the original and sparsified 3D maps consisting of reference images and 3D points. The original map contains a lot of redundant images and 3D points, resulting in a huge map size. By defining 3D landmarks and assigning each landmark a virtual reference frame, the number of images and 3D points are significantly reduced, leading to a more compact map. For example, the original map has 744 reference images and 28.3k 3D points in apt1/kitchen of 12Scenes dataset~\cite{twelvescenes2016}, while \method has only 16 virtual reference images and 10.0k 3D points. In StMarys Church of CambridgeLandmarks~\cite{posenet}, the original map has 1,487 reference frames and 379.8k 3D points, while the numbers are reduced to 32 and 134.8k. On the Aachen dataset~\cite{aachen}, PRAM discards over 90\% of reference images (479 vs. 6,697) and 40\% of 3D points (1.1m vs. 1.8m), significantly reducing the map size.

\begin{figure}[t]
	\centering
	\includegraphics[width=1.0\linewidth]{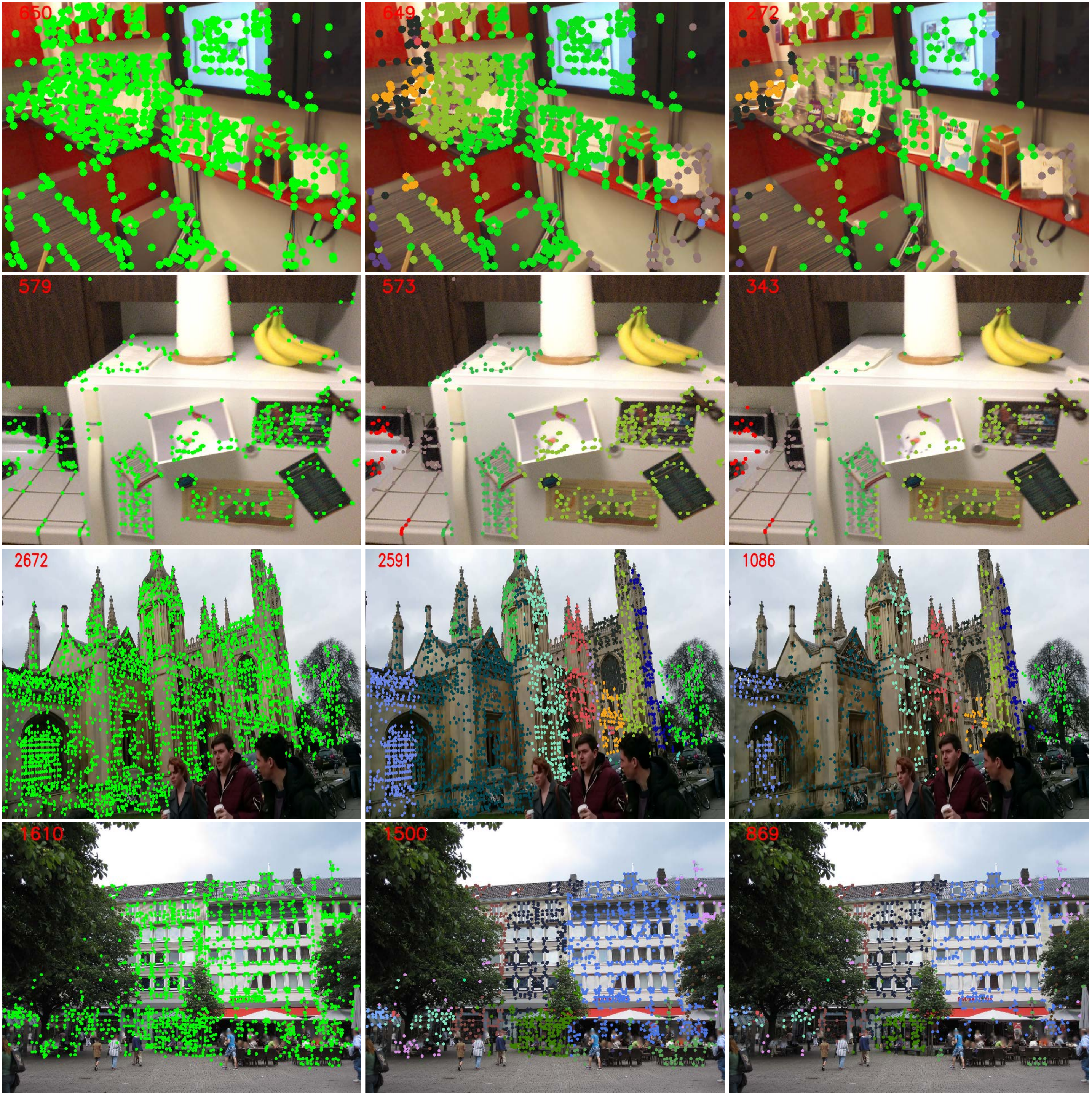}
	\caption[Visualization of sparsification on 2D keypoints]{\textbf{Sparsification of 2D keypoints.} Original keypoints (left), keypoints with spatial consistency (middle), and finally sparsified keypoints (right) of images from 7Scenes~\cite{sevenscenes2013}, 12Scenes~\cite{twelvescenes2016}, Cambridgelandmarks~\cite{posenet}, and Aachen~\cite{aachen} datasets are visualized (from top to bottom).}
	\label{fig:sparse_points}
\end{figure}

\textbf{Keypoint sparsification.} Fig.~\ref{fig:sparse_points} shows the sparsification of redundant 2D points on images from 7Scenes~\cite{sevenscenes2013}, 12Scenes~\cite{twelvescenes2016}, CambridgeLandmarks~\cite{posenet}, and Aachen~\cite{aachen} datasets. The original points are those with 3D correspondences. Due to insufficient observations, 3D points with large noise are removed as introduced in Sec.~\ref{sec:method:landmark_definition}. Therefore, some points from dynamic objects, such as pedestrians (row 3), and tables (row 4) are filtered. After filtering, existing keypoints still have high overlap, so our adaptive pruning strategy further discards redundant ones according to their overlap. After pruning, the number of points for both indoor (row 1 and 2) and outdoor (row 3 and 4) scenes is significantly reduced by even up to 50\%. The sparsification does not cause an obvious performance drop as discussed in Sec.~\ref{sec:exp:ablation}.

\begin{figure}[t]
	\centering
	\includegraphics[width=1.\linewidth]{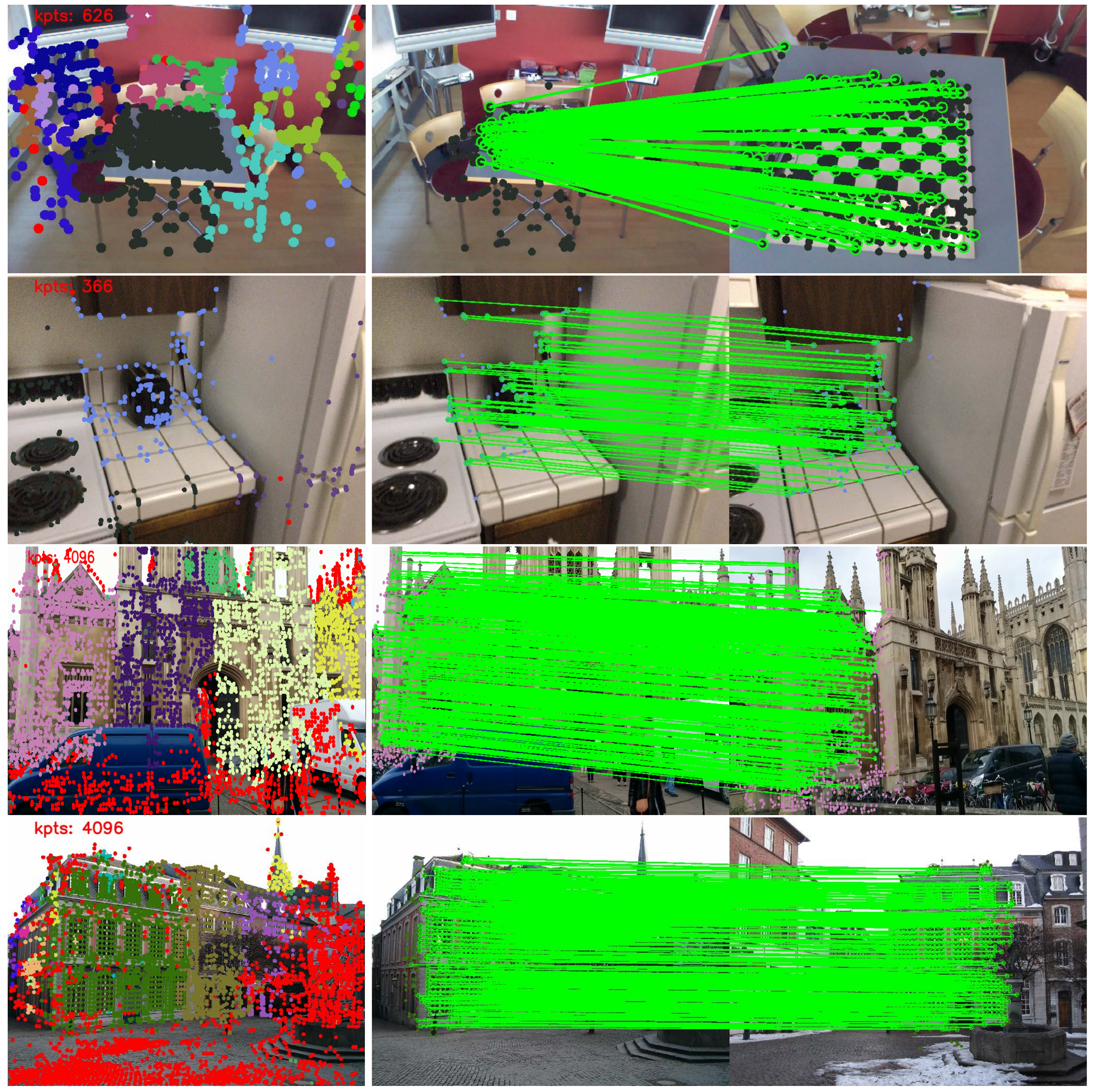}
	\caption[Landmark-wise matching]{\textbf{Outlier identification and landmark-wise matching.} The left shows the predicted landmark labels of keypoints including those identified as outliers (\textcolor{red}{red points}) of query images. These outliers can be easily removed without using any manually defined tricks. The right presents the matches between query and virtual reference frames. With the guidance of the self-defined landmarks, fast landmark-wise matching rather than exhaustive matching is executed. From top to bottom, results on 7Scenes~\cite{sevenscenes2013}, 12Scenes~\cite{twelvescenes2016}, CambridgeLandmarks~\cite{posenet}, and Aachen~\cite{aachen} datasets are included.}
	\label{fig:landmark_matching}
\end{figure}

\textbf{Outlier identification and landmark-wise matching.} Fig.~\ref{fig:landmark_matching} showcases the outlier recognition and landmark-wise matching at test time. Fig.~\ref{fig:landmark_matching} (left) visualizes the potential outliers (\textcolor{red}{red points}) which are mainly from dynamic objects and trees. As these keypoints do not have potential correspondences in the map, they can be discriminated easily by our recognition module instead of relying on manually-defined tricks. The landmark-wise matching (right) is conducted between the query and virtual reference images to reduce the domain differences between 2D and 3D points. Fig.~\ref{fig:landmark_matching} shows that the self-defined landmarks provide strong guidance for matching. These landmarks work well in both indoor and outdoor scenes with high generalization ability.

\section{Conclusion, Limitation, And Future Work}
\label{sec:conclusion_futurework}

In this section, we conclude this paper in Sec.~\ref{sec:conclution} and discuss the limitations and future works in Sec.~\ref{sec:futurework}

\subsection{Conclusion}
\label{sec:conclution}

In this paper, we propose the place recognition anywhere model (PRAM) for efficient and accurate localization in large-scale environments. Specifically, we adopt a self-supervised strategy to generate landmarks in the 3D space directly in a self-supervised manner. This strategy allows PRAM to make any place a landmark without the limitation of commonly used classic semantic labels. By representing the map with 3D landmarks, PRAM discards global descriptors, repetitive local descriptors, and redundant 3D points and thus effectively reduces the map size. Additionally, PRAM performs visual localization efficiently by recognizing landmarks from sparse keypoints instead of dense pixels. The recognized landmarks from sparse keypoints are further used for outlier removal and landmark-wise 2D-3D matching rather than exhaustive 2D-2D matching, leading to higher time efficiency. A comprehensive evaluation of previous APRs, SCRs, HMs, and PRAM on public indoor 7Scenes, 12Scenes, and outdoor CambridgeLandmarks and Aachen datasets demonstrates that PRAM outperforms ARPs and SCRs in large-scale environments and gives competitive accuracy to HMs, but has higher memory and time efficiency, paving a way towards efficient and accurate visual localization.

\subsection{Limitation and future work}
\label{sec:futurework}

\method, as a new localization framework, achieves a better balance between efficiency and accuracy compared with APRs, SCRs, and HMs in large-scale environments. However, in outdoor scenes, \method still has about 7\% lower accuracy than HMs. In this section, we discuss potential reasons and corresponding solutions.

\textbf{Landmark generation.} The current strategy of generating landmarks only considers the spatial connections of 3D points. The object-level consistency is ignored, which may lead to the split of objects, impairing the recognition and localization accuracy. In the future, object masks given by SAM~\cite{sam2023} can be used as additional object-level connections in the hierarchical clustering process to mitigate this problem. The noise of SAM results especially on outdoor scenes should be reduced to avoid the inconsistency between multi-view images. In addition, the number of landmarks for each dataset is manually set according to their spatial sizes, which may not be optimal. An adaptive solution to determining the number of landmarks for a scene is worth further exploration. 

\textbf{Map compression.} How to reduce map size remains a challenge in hierarchical frameworks. By representing the map hierarchically with landmarks, PRAM provides a new way of map compression. However, PRAM still needs to store descriptors for each 3D point to guarantee its accuracy (as discussed in Sec.~\ref{sec:exp:ablation}). In the future, these explicit 3D points and their descriptors in each 3D landmark can be replaced with a more compact tiny NeRF model~\cite{nerfs2021}. Meanwhile, in contrast to using one of the reference images in the database as the virtual reference frame, the viewpoint of the virtual reference frame for each landmark can be rendered with NeRFs to better describe this landmark.


\textbf{Multi-modality localization.} One advantage of PRAM is leveraging sparse keypoints as tokens to conduct landmark recognition, making visual localization not only a low-level geometric task but also a high-level perception task. However, this also leads to a slight loss of accuracy compared with HMs. As sparse keypoints sometimes are not robust to dynamic objects and texture-less regions, other visual features including lines~\cite{lineflow,gluestick2023} and multi-modality signals such as GPS, Magnetometer signals, texts, and audios could be incorporated to enhance the recognition. By making localization a high-level perception task, PRAM provides an easy way of using these signals. For example, these signals can be encoded as tokens to propagate complementary knowledge to sparse keypoints via the cross-attention mechanism. 





\ifCLASSOPTIONcompsoc
\section*{Acknowledgments}
\else
\section*{Acknowledgment}
\fi

This project is supported by Toyota Motor Europe.

\bibliographystyle{IEEEtran}
\bibliography{egbib}

\begin{IEEEbiography}[{\includegraphics[width=1in,height=1.25in,clip,keepaspectratio]{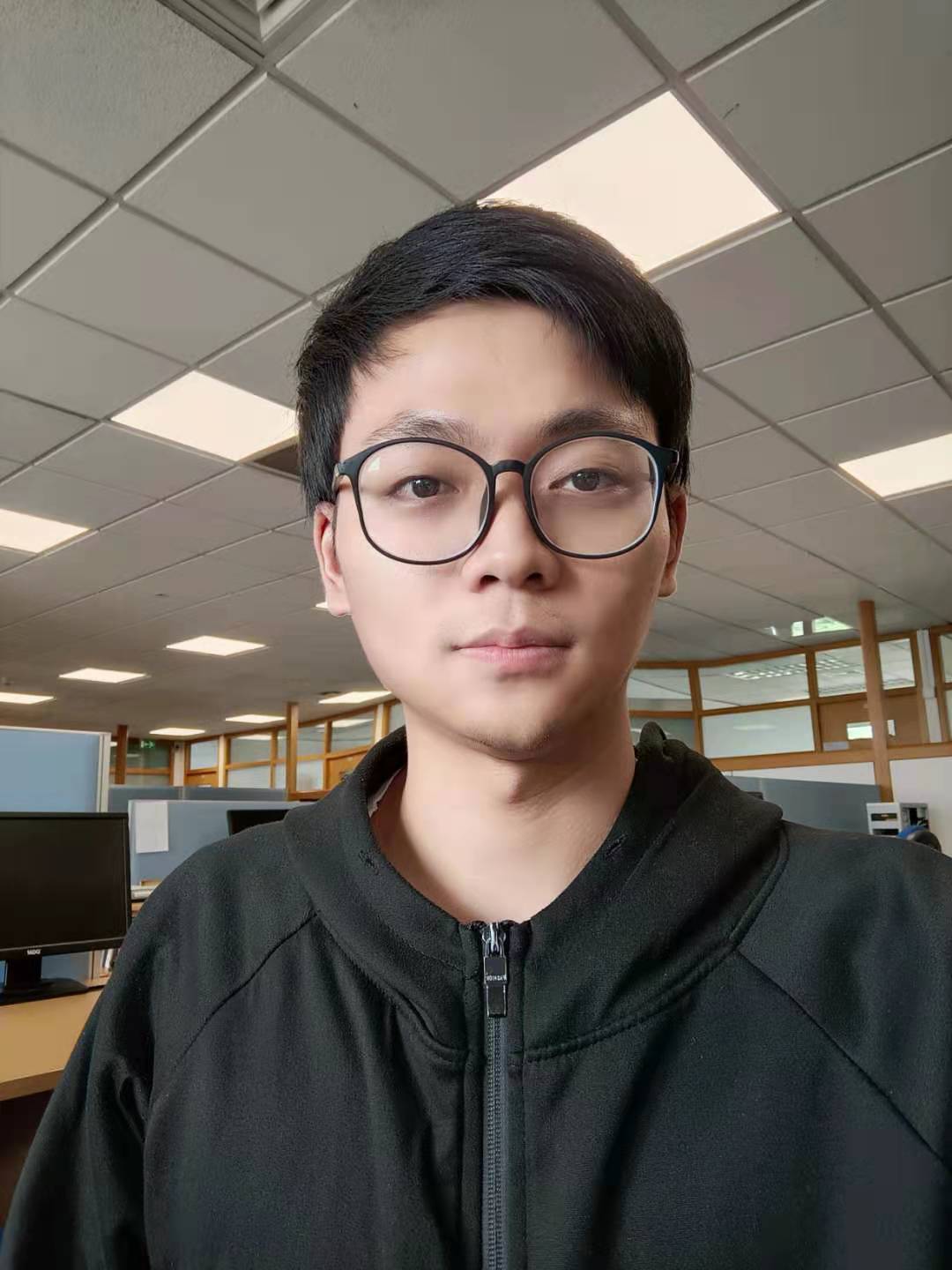}}]{Fei Xue}
	received his B.E. in 2016 from School of Electronics Engineering and Computer Science, Peking University, Beijing, China, 100871. In the same year, he joined the Key Laboratory of Machine Perception (Ministry of Education) as a master student and graduated in the year of 2019. He is now a PhD student in Machine Intelligence Laboratory, Department of Engineering, University of Cambridge. His interests include visual simultaneous localization and mapping (SLAM), visual relocalization, 3D semantic reconstruction, feature extraction and matching. He is a member of the IEEE.
\end{IEEEbiography}

\begin{IEEEbiography}[{\includegraphics[width=1in,height=1.25in,clip,keepaspectratio]{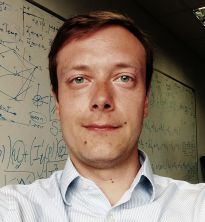}}]{Ignas Budvytis}
	received the BA degree in computer science from the University of Cambridge in 2008. He received his doctoral studies in the Machine Intelligence Laboratory, Department of Engineering, University of Cambridge in 2013. He is now an assistant professor in the Machine Intelligence Laboratory, Department of Engineering, University of Cambridge. His research interests include semi-supervised video segmentation, human body reconstruction and pose estimation, visual localization, and object class recognition. He is a member of the IEEE.
\end{IEEEbiography}

\begin{IEEEbiography}[{\includegraphics[width=1in,height=1.25in,clip,keepaspectratio]{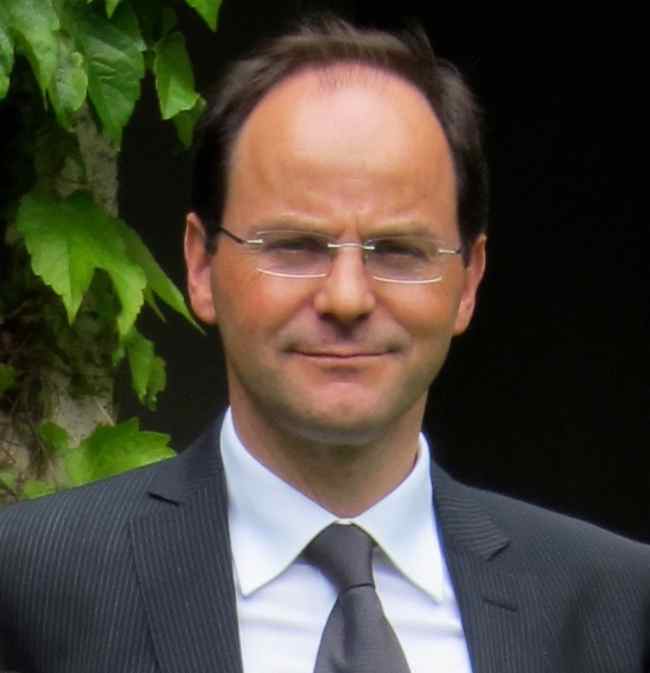}}]{Roberto Cipolla} 
	received the BA degree in engineering from the University of Cambridge, in 1984, the MSE degree in electrical engineering
	from the University of Pennsylvania, in 1985, and
	the DPhil degree in computer vision from the University of Oxford, in 1991. From 1991-92 he was a Toshiba fellow and engineer in the Toshiba Corporation Research and Development Center, Kawasaki, Japan. He joined the Department of Engineering, University of Cambridge, in 1992 as a lecturer and a fellow of Jesus College. He
	became a reader in information engineering in 1997 and a professor in
	2000. He became a fellow of the Royal Academy of Engineering
	(FREng), in 2010 and a fellow of the Royal Society (FRS), in 2022. His research interests include computer vision and
	robotics. He has authored 3 books, edited 9 volumes and co-authored
	more than 300 papers. He is a senior member of the IEEE.
\end{IEEEbiography}

\end{document}